%% file: PaperForReview.tex
\documentclass[10pt,twocolumn,letterpaper]{article}

\usepackage[pagenumbers]{cvpr} 

\usepackage{graphicx}
\usepackage{amsmath}
\usepackage{amssymb}
\usepackage{booktabs}

\usepackage{bm}
\usepackage{bbding}
\usepackage{multirow}
\usepackage{color}

\usepackage{pifont}
\newcommand{\cmark}{\ding{51}}%
\newcommand{\xmark}{\ding{55}}%

\usepackage[pagebackref,breaklinks,colorlinks]{hyperref}

\usepackage[capitalize]{cleveref}
\crefname{section}{Sec.}{Secs.}
\Crefname{section}{Section}{Sections}
\Crefname{table}{Table}{Tables}
\crefname{table}{Tab.}{Tabs.}

\def\cvprPaperID{9674} 
\def\confName{CVPR}
\def\confYear{2022}

\newenvironment{shrinkeq}[1]
{\bgroup
  \addtolength\abovedisplayshortskip{#1}
  \addtolength\abovedisplayskip{#1}
  \addtolength\belowdisplayshortskip{#1}
  \addtolength\belowdisplayskip{#1}}
{\egroup\ignorespacesafterend}

\begin{document}

\title{AdaFocus V2: End-to-End Training \\ 
of Spatial Dynamic Networks for Video Recognition}

\author{
Yulin Wang$^1$\thanks{Equal contribution.\ \ \ \ \ \ \ \ \ \ \ \ \ \ \ \textsuperscript{\Envelope}Corresponding authors.} \ \ \ \ \ \ \ \ 
Yang Yue$^1$$^*$ \ \ \ \ \ \ \ \ \ \ 
Yuanze Lin$^2$ \ \ \ \ \ \ \ \ \ \ 
Haojun Jiang$^1$ \ \ \ \ \ \ \ \ \ \ 
Zihang Lai$^3$\\
Victor Kulikov$^4$ \ \ \ \ \ \ \ \ \ \ 
Nikita Orlov$^4$\ \ \ \ \ \ \ \ \ \ 
Humphrey Shi$^{4, 5}$ \!\textsuperscript{\Envelope}\ \ \ \ \ \ \ \ \ \ 
Gao Huang$^{1}$ \!\textsuperscript{\Envelope}\\
{\small$^1$Department of Automation, BNRist, Tsinghua University\ \ \ \ \ \ $^2$U of Washington\ \ \ \ \ \ $^3$CMU}\\
{\small$^4$Picsart AI Research (PAIR) \ \ \ \ \ \ $^5$U of Oregon}\\
}

\maketitle

\begin{abstract}
  Recent works have shown that the computational efficiency of video recognition can be significantly improved by reducing the spatial redundancy. As a representative work, the adaptive focus method (\textbf{AdaFocus}) has achieved a favorable trade-off between accuracy and inference speed by dynamically identifying and attending to the informative regions in each video frame. However, AdaFocus requires a complicated three-stage training pipeline (involving reinforcement learning), leading to slow convergence and is unfriendly to practitioners. This work reformulates the training of AdaFocus as a simple one-stage algorithm by introducing a differentiable interpolation-based patch selection operation, enabling efficient end-to-end optimization. We further present an improved training scheme to address the issues introduced by the one-stage formulation, including the lack of supervision, input diversity and training stability. Moreover, a conditional-exit technique is proposed to perform temporal adaptive computation on top of AdaFocus without additional training. Extensive experiments on six benchmark datasets (i.e., ActivityNet, FCVID, Mini-Kinetics, Something-Something V1\&V2, and Jester) demonstrate that our model significantly outperforms the original AdaFocus and other competitive baselines, while being considerably more simple and efficient to train. Code is available at \url{https://github.com/LeapLabTHU/AdaFocusV2}.

\end{abstract}


\input{introduction.tex}
\input{related.tex}

\input{method.tex}

\input{experiment.tex}

\input{conclusion.tex}

\section*{Acknowledgements}
This work is supported in part by the National Science and Technology Major Project of the Ministry of Science and Technology of China under Grants 2018AAA0100701, the National Natural Science Foundation of China under Grants 61906106 and 62022048, Picsart AI Research and Beijing Academy of Artificial Intelligence. 

{\small
\bibliographystyle{ieee_fullname}
\bibliography{egbib}
}

\section*{Appendix}

\appendix


\section{Datasets and Baselines}

\subsection{Datasets}

Our experiments are based on six widely-used large-scale video recognition datasets. For all of them, we use the official training-validation split.
\begin{itemize}
  \item ActivityNet \cite{caba2015activitynet} contains 10,024 training videos and 4,926 validation videos sorted into 200 human action categories. The average duration is 117 seconds.
  \item FCVID \cite{TPAMI-fcvid} contains 45,611 videos for training and 45,612 videos for validation, which are annotated into 239 classes. The average duration is 167 seconds.
  \item Mini-Kinetics is a subset of the Kinetics \cite{kay2017kinetics} dataset. We establish it following \cite{wu2019liteeval, meng2020ar, meng2021adafuse, sun2021dynamic, kim2021efficient, ghodrati2021frameexit}. The dataset include 200 classes of videos, 121k for training and 10k for validation. The average duration is around 10 seconds \cite{kay2017kinetics}.
  \item Something-Something (Sth-Sth) V1\&V2 \cite{goyal2017something} datasets include 98k and 194k videos respectively. Both of them are labeled with 174 human action classes. The average duration is 4.03 seconds.
  \item Jester \cite{materzynska2019jester} dataset consists of 148,092 videos in 27 action categories. The average duration is 3 seconds.
\end{itemize}

\textbf{Data pre-processing.}
Following \cite{lin2019tsm, meng2020ar,Wang_2021_ICCV}, the training data is augmented via random scaling followed by 224x224 random cropping, after which random flipping is performed on all datasets except for Sth-Sth V1\&V2 and Jester. At test time, since we consider improving the inference efficiency of video recognition, we resize the short side of video frames to 256 and perform 224x224 centre-crop, obtaining a single clip per video for evaluation.

\subsection{Baselines}
In addition to AdaFocusV1, our proposed AdaFocusV2 approach is compared with several state-of-the-art frameworks designed for efficient video recognition, including MultiAgent \cite{wu2019multi}, LiteEval \cite{wu2019liteeval}, SCSampler \cite{korbar2019scsampler}, ListenToLook \cite{gao2020listen}, AR-Net \cite{meng2020ar}, AdaFrame \cite{wu2019adaframe}, AdaFuse \cite{meng2021adafuse}, VideoIQ \cite{sun2021dynamic}, Dynamic-STE \cite{kim2021efficient} and FrameExit \cite{ghodrati2021frameexit}. Here we briefly introduce them.
\begin{itemize}
  \item MultiAgent \cite{wu2019multi} learns to attend to important frames using multi-agent reinforcement learning. The implementation in \cite{meng2020ar} is adopted.
  \item LiteEval \cite{wu2019liteeval} allocates computation dynamically according to the importance of frames by switching between coarse and fine LSTM networks.
  \item SCSampler \cite{korbar2019scsampler} is an efficient framework to select salient temporal clips from a long video. The implementation in \cite{meng2020ar} is adopted.
  \item ListenToLook \cite{gao2020listen} selects the key clips of a video by leveraging audio information. We adopt the image-based variant introduced in their paper for fair comparisons, since we do not use the audio of videos.
  \item AR-Net \cite{meng2020ar} processes video frames with different resolutions based on their relative importance.
  \item AdaFrame \cite{wu2019adaframe} learns to adaptively identify informative frames on a per-video basis with reinforcement learning. Each video is processed using different numbers of frames, facilitating dynamic inference.
  \item AdaFuse \cite{meng2021adafuse} proposes to dynamically fuse channels along the temporal dimension for modeling temporal relationships effectively.
  \item VideoIQ \cite{sun2021dynamic} learns to select optimal precision for each frame conditioned on their importance in terms of video recognition. 
  \item Dynamic-STE \cite{kim2021efficient} adopts a lighter student network and a heavier teacher network to process more and less frames, respectively. The two networks dynamically interact with each other during inference.
  \item FrameExit \cite{ghodrati2021frameexit} learns to process relatively fewer frames for simpler videos and more frames for difficult ones.
\end{itemize}

\section{Implementation Details}

\subsection{Architecture of the Policy Network $\pi$}
We follow the design of $\pi$ adopted by AdaFocusV1 \cite{Wang_2021_ICCV}. The global feature maps $\bm{e}^{\textnormal{G}}_{t}$ of each frame is compressed to 64 channels by a 1x1 convolutional layer, vectorized, and fed into a one-layer gated recurrent unit (GRU) \cite{cho-etal-2014-learning} with a hidden size of 2048. The outputs are projected to 2 dimensions (i.e., $(\tilde{x}^t_{\textnormal{c}}, \tilde{y}^t_{\textnormal{c}})$) and processed by the sigmoid activation function. On top of TSM, since a single patch location is generated for each video, we concatenate $\bm{e}^{\textnormal{G}}_{t}$ of all frames as the input of $\pi$, and replace the GRU by an MLP.

\subsection{Training Hyper-parameters}
In general, we find that the performance of AdaFocusV2 does not rely on the extensive hyper-parameter searching on a dataset or patch size basis. The training hyper-parameters only need to be tuned when the backbone networks (i.e., $f_{\textnormal{G}}$ and $f_{\textnormal{L}}$) change, and it may largely follow the training protocol for solely training the backbones (e.g., typically, this can be easily obtained from the official implementation in the literature).

\textbf{ActivityNet, FCVID and Mini-Kinetics (Section 4.1).}
As stated in the paper, all the components (i.e., $f_{\textnormal{G}}$, $f_{\textnormal{L}}$, $f_{\textnormal{C}}$ and $\pi$) of AdaFocusV2 are trained simultaneously in a standard end-to-end fashion. An SGD optimizer with cosine learning rate annealing and a momentum of 0.9 is adopted. The L2 regularization co-efficient is set to 1e-4. The two encoders $f_{\textnormal{G}}$ and $f_{\textnormal{L}}$ are initialized using the ImageNet pre-trained models\footnote{In most cases, we use the 224x224 ImageNet pre-trained models provided by PyTorch \cite{paszke2019pytorch}. In AdaFocusV2-RN, since we deploy a ResNet-50 with down-sampled inputs ($96^2$) as $f_{\textnormal{G}}$, we use the 96x96 ImageNet pre-trained ResNet-50 (provided by \cite{NeurIPS2020_7866}).} , while $f_{\textnormal{C}}$ and $\pi$ are trained from random initialization. On ActivityNet and FCVID, the size of the mini-batch is set to 32. The initial learning rates of $f_{\textnormal{G}}$, $f_{\textnormal{L}}$, $f_{\textnormal{C}}$ and $\pi$ are set to 0.001, 0.002, 0.01 and 2e-4, respectively. On Mini-Kinetics, we adopt a batch size of 48, and linearly scale the initial learning rates. The experiments with all patch sizes use the same aforementioned training configurations.


\textbf{Sth-Sth V1\&V2 and Jester (Section 4.2).}
When TSM \cite{lin2019tsm} is implemented as the backbones in AdaFocusV2, the initial learning rates of $f_{\textnormal{G}}$, $f_{\textnormal{L}}$, $f_{\textnormal{C}}$ and $\pi$ are set to 0.005, 0.01, 0.01 and 1e-4, respectively. The L2 regularization co-efficient is set to 5e-4. These changes follow the official implementation of TSM \cite{lin2019tsm}. All other training settings are the same as the experiments on ActivityNet/FCVID in Section 4.1. All the experiments on Sth-Sth V1\&V2 and Jester adopt the same training configurations.

\end{document}

%% file: introduction.tex
\vspace{-1ex}
\section{Introduction}
\vspace{-1ex}

\begin{table}[t]
  \centering
  \begin{footnotesize}
  \caption{\textbf{A comparison of training the original AdaFocus model (AdaFocusV1, from \ding{182} to \ding{186}) and AdaFocusV2 (end-to-end) on Something-Something (Sth-Sth) V1 dataset.} Both procedures start from the same initial backbone networks. Herein, $f_{\textnormal{G}}$, $f_{\textnormal{L}}$, $f_{\textnormal{C}}$ and $\pi$ are the model components (see Section \ref{sec:31} for details).}
  \vskip -0.175in
  \setlength{\tabcolsep}{3mm}{
  \vspace{5pt}
  \renewcommand\arraystretch{0.95}
  \resizebox{0.9\columnwidth}{!}{
  \begin{tabular}{cl|c}
  \toprule
   \multicolumn{2}{c|}{AdaFocusV1} & AdaFocusV2 \\
   \midrule
   \midrule
   \multirow{2}{*}{\textbf{\emph{Pre-training}}} & \ding{182} Pre-train $f_{\textnormal{G}}$ on Sth-Sth V1.& \multirow{8.5}{*}{\shortstack{\textcolor{blue}{{\textbf{\emph{End-to-End}}}}\\ \textbf{\emph{\textcolor{blue}{Training}}}\\ \textcolor{black}{{\emph{($f_{\textnormal{G}}$, $f_{\textnormal{L}}$, $f_{\textnormal{C}}$, $\pi$)}}}}} \\
    & \ding{183} Pre-train $f_{\textnormal{L}}$ on Sth-Sth V1.& \\
  \cmidrule{1-2}
   \multirow{2}{*}{\textbf{\emph{Stage-1}}} & \ding{184} Train $f_{\textnormal{L}}$ and $f_{\textnormal{C}}$ using & \\
    &random patches.& \\
    \cmidrule{1-2}
    \multirow{2}{*}{\textbf{\emph{Stage-2}}} & \ding{185} Train $\pi$ using & \\
    &reinforcement learning.& \\
    \cmidrule{1-2}
    {\textbf{\emph{Stage-3}}} & \ding{186} Fine-tune $f_{\textnormal{L}}$ and $f_{\textnormal{C}}$. & \\
  \bottomrule
  \label{tab:tab1}
  \end{tabular}}}
  \end{footnotesize}
  \vskip -0.25in
\end{table}

\begin{figure}[t]
  \begin{center}
  \centerline{\includegraphics[width=0.8\columnwidth]{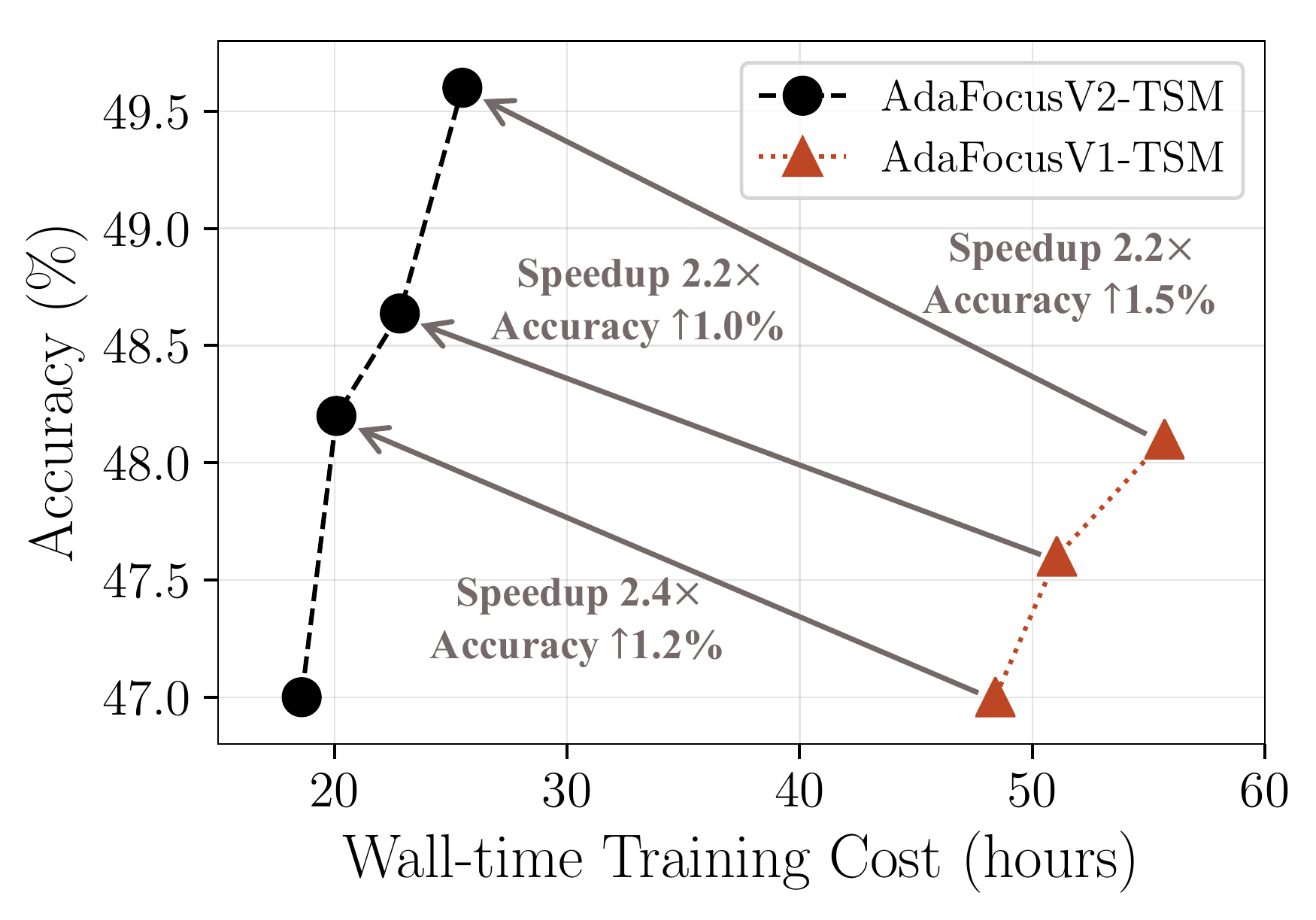}}
  \vskip -0.15in
  \caption{\textbf{Comparisons of AdaFocusV1 and AdaFocusV2 on Sth-Sth V1 in terms of accuracy v.s. training cost.} The training time is measured based on 4 NVIDIA 3090 GPUs. The two sides of grey arrows correspond to the same network architecture (i.e., the same inference cost). Our AdaFocusV2 accelerates the training by \textbf{2.2-2.4$\bm{\times}$}, while boosting the accuracy by \textbf{1.0-1.5$\bm{\%}$}. \label{fig:fig1}
  }
  \end{center}
  \vskip -0.4in
\end{figure}

Deep networks have achieved remarkable success in large-scale video recognition tasks \cite{tran2015learning, feichtenhofer2016convolutional, carreira2017quo, hara2018can, feichtenhofer2019slowfast}. Their high accuracy has fueled the desire to deploy them for automatically recognizing the actions, events, or other contents within the explosively growing online videos in recent years (e.g., on YouTube). However, the models with state-of-the-art performance \cite{ghadiyaram2019large, tran2019video, qiu2019learning, feichtenhofer2020x3d, liu2021video, arnab2021vivit} tend to be computationally intensive during inference. In real-world applications such as recommendation \cite{davidson2010youtube, deldjoo2016content, gao2017unified}, surveillance \cite{collins2000system, chen2019distributed} and content-based searching \cite{ikizler2007searching}, computation translates into power consumption and latency, both of which should be minimized for environmental, safety or economic reasons.

Several algorithms have been proposed to reduce the \emph{temporal redundancy} of videos \cite{yeung2016end, wu2019adaframe, gao2020listen, korbar2019scsampler, wu2019multi, meng2020ar, ghodrati2021frameexit, kim2021efficient, sun2021dynamic} by allocating the majority of computation to the most task-relevant video frames rather than all. Orthogonal to these approaches, the recently proposed adaptive focus network (AdaFocus) \cite{Wang_2021_ICCV} reveals that reducing the \emph{spatial redundancy} in video analysis yields promising results for efficient video recognition. 
The AdaFocus framework is also compatible with the aforementioned temporal-adaptive methods to realize highly efficient spatial-temporal computation. 
Specifically, AdaFocus reduces the computational cost by applying the expensive high-capacity network only on some relatively small patches. These patches are strategically selected to capture the most informative regions of each video frame. 
In particular, the patch localization task is formulated as a non-differentiable \emph{discrete} decision task, which is further solved with reinforcement learning. 
As a consequence, AdaFocus needs to be trained with a complicated three-stage training pipeline (see Table~\ref{tab:tab1}), resulting in long training time and being unfriendly to users.

This paper seeks to simplify the training process of AdaFocus. We first introduce a \emph{differentiable} interpolation-based formulation for patch selecting, allowing gradient back-propagation throughout the whole model. We note that a straightforward implementation of end-to-end training leads to optimization issues, including the lack of supervision, input diversity and training stability, which severely degrade the performance. Therefore, we further propose three tailored training techniques: auxiliary supervision, diversity augmentation and stopping gradient, to address the aforementioned issues. These simple but effective techniques enable the simple one-stage formulation of our AdaFocus algorithm to be trained effectively, and eventually outperform the three-stage counterparts in terms of both test accuracy and training cost. The experimental comparisons are presented in Figure \ref{fig:fig1}. Our proposed method is referred to as AdaFocusV2.

An additional advantage of AdaFocus is that it can be easily improved by further considering \emph{temporal redundancy}. The original paper implements this idea by dynamically skipping less valuable frames with reinforcement learning. In contrast, this work proposes a simplified early-exit algorithm that removes the requirement of introducing additional training, but achieves competitive performance. 


The effectiveness of AdaFocusV2 is extensively evaluated on six video recognition benchmarks (i.e., ActivityNet, FCVID, Mini-Kinetics, Something-Something V1\&V2, and Jester). Experimental results show that the training of AdaFocusV2 is 2$\times$ faster (measured in wall-time) than the original counterpart, while achieving consistently higher accuracy.

%% file: related.tex
\vspace{-0.5ex}
\section{Related Works}
\vspace{-0.5ex}

\textbf{Video recognition.}
In recent years, convolutional networks (ConvNets) have achieved remarkable performance for video recognition. One of the prominent approaches is to capture the spatial/temporal information jointly using 3D ConvNets. Representative works include C3D \cite{tran2015learning}, I3D \cite{carreira2017quo}, ResNet3D \cite{hara2018can}, X3D \cite{feichtenhofer2020x3d}, etc. Some other works focus on first extracting frame-wise features, and then aggregating temporal information with specialized architectures, such as temporal averaging \cite{wang2016temporal}, deploying recurrent networks \cite{donahue2015long, li2018recurrent, yue2015beyond}, and temporal channel shift \cite{lin2019tsm, sudhakaran2020gate, fan2020rubiksnet, meng2021adafuse}. Another line of works leverage two-stream architectures to model short-term and long-term temporal relationships respectively \cite{feichtenhofer2016convolutional, feichtenhofer2017spatiotemporal, feichtenhofer2019slowfast, gong2021searching}. In addition, as processing videos with ConvNets, especially 3D ConvNets, tends to be computationally intensive, recent research starts to pay attention to designing efficient video recognition models \cite{tran2018closer, zolfaghari2018eco, tran2019video, luo2019grouped, liu2020teinet, liu2021tam}.

\textbf{Temporal redundancy.}
A popular approach for facilitating efficient video recognition is to reduce the temporal redundancy in videos \cite{yeung2016end, wu2019adaframe, gao2020listen, korbar2019scsampler, wu2019multi, meng2020ar, ghodrati2021frameexit, kim2021efficient, sun2021dynamic}. Since not all frames are equally important for a given task, the model should ideally allocate less computation on less informative frames \cite{han2021dynamic}. Several effective algorithms have been proposed in this direction. For example, VideoIQ \cite{sun2021dynamic} processes video frames using different precision according to their relative importance. FrameExit \cite{ghodrati2021frameexit} learns to terminate the inference process after seeing a few sufficiently informative frames. The AdaFocusV2+ algorithm, which we propose to model temporal redundancy on top of AdaFocusV2, is related to FrameExit on the spirit of early-exit. However, our method is easier to implement since it does not need to learn an additional conditional-exit policy.


\textbf{Spatial-wise dynamic networks}
perform computation adaptively on top of different spatial locations of the inputs \cite{han2021dynamic, jaderberg2015spatial}. The AdaFocusV2 network studied in this paper can be classified into this category as well. Many of the spatially adaptive networks are designed from the lens of inference efficiency \cite{han2021dynamic, ren2018sbnet, yang2020resolution, wang2019adaptively, chen2021dynamic}. For example, recent works have revealed that 2D images can be efficiently processed via attending to the task-relevant or more informative image regions \cite{figurnov2017spatially, NeurIPS2020_7866, xie2020spatially, verelst2020dynamic}. In the context of video recognition, how to exploit this spatial redundancy for reducing the computational cost is still an under-explored topic. It has been shown by the attention-based methods \cite{meng2019interpretable} that the contributions of different frame regions to the recognition task are not equivalent. Preliminary works like AdaFocus \cite{han2021dynamic} have demonstrated the potentials of this direction.

The spatial transformer networks \cite{jaderberg2015spatial} are trained based on a similar interpolation-based mechanism to us. However, they focus on actively transforming the feature maps for learning spatially invariant representations, while we aim to localize and attend to the task-relevant regions of the video frames for improving the inference efficiency. Moreover, we show that a straightforward implementation of this mechanism fails to yield competitive results in our problem. Special designs need to be introduced by our algorithm to solve the optimization difficulties. 



%% file: method.tex
\vspace{-0.5ex}
\section{Method}
\vspace{-0.5ex}


The AdaFocus network \cite{Wang_2021_ICCV} facilitates efficient video recognition via reducing the spatial-wise redundant computation. However, it suffers from a complicated three-stage training procedure. This section introduces the details of our end-to-end trainable AdaFocusV2 approach, which consistently outperforms the original AdaFocus with a simpler and more efficient training procedure.

\subsection{Preliminaries of AdaFocusV1 \cite{Wang_2021_ICCV}}
\label{sec:31}

\begin{figure}[t]
    \begin{center}
    \centerline{\includegraphics[width=\columnwidth]{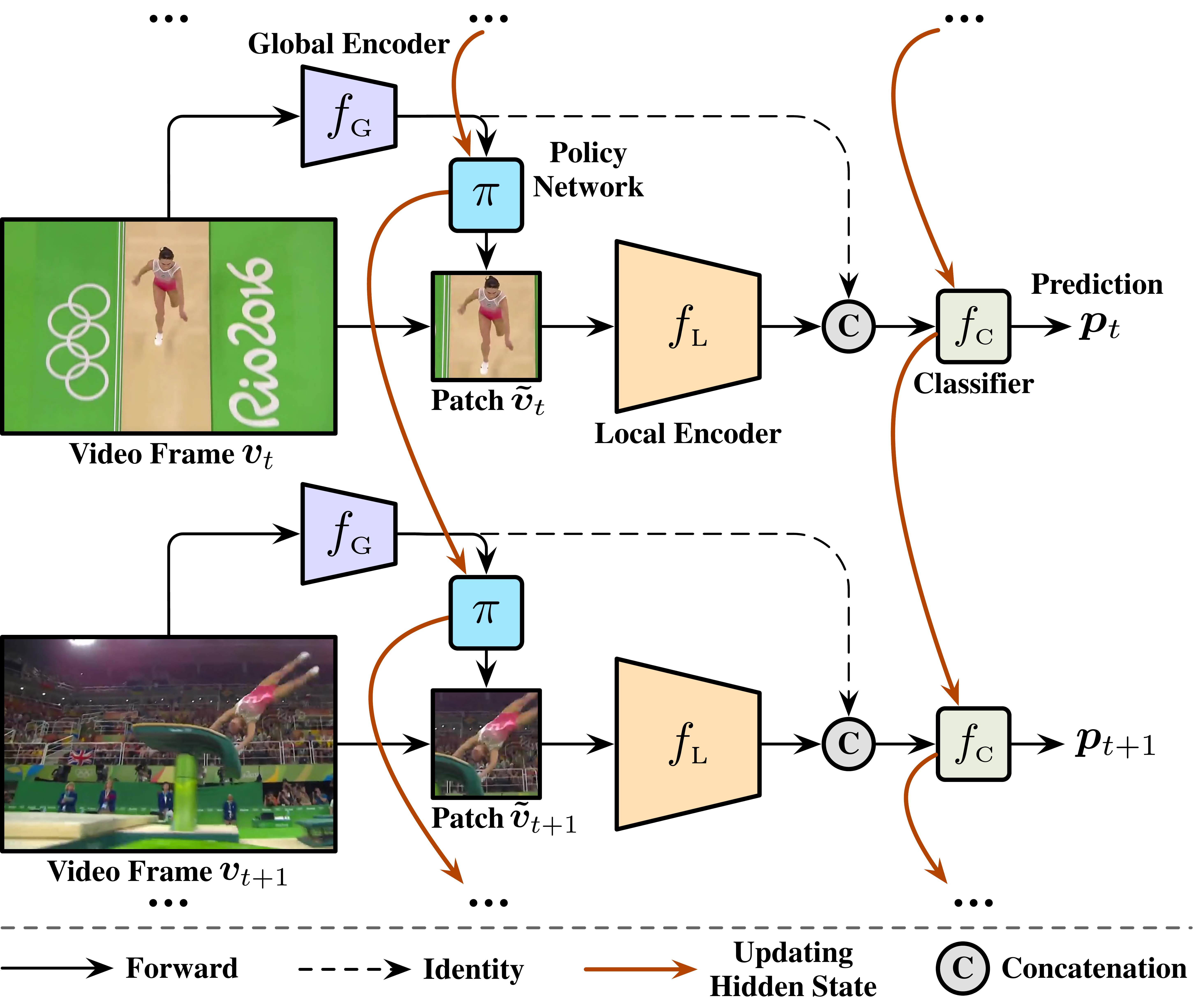}}
    \caption{\textbf{An overview of adaptive focus networks (AdaFocus).} 
    The global and local encoders $f_{\textnormal{G}}$ and $f_{\textnormal{L}}$ are two deep networks (e.g., ConvNets). The former catches a glimpse of each frame, enabling the recurrent policy network $\pi$ to localize an informative image patch $\tilde{\bm{v}}_t$. The latter is designed to extract discriminative representations by processing $\tilde{\bm{v}}_t$ with its large and high-capacity architecture. A classifier $f_{\textnormal{C}}$ aggregates the information from all the processed frames to obtain the prediction. 
    \label{fig:overview}
    }
    \end{center}
    \vskip -0.25in
\end{figure}

We start by giving an overview of AdaFocus (see Figure \ref{fig:overview}), laying the basis for the discussions on training efficiency. Assume that a stream of video frames $\{\bm{v}_1, \bm{v}_2, \ldots\}$ comes in sequentially. AdaFocus first takes a quick glance at each frame with a light-weighted global encoder $f_{\textnormal{G}}$, aiming to extract cheap and coarse global features
\begin{shrinkeq}{-0ex}
\begin{equation}
    \bm{e}^{\textnormal{G}}_{t} = f_{\textnormal{G}}(\bm{v}_t),\ \ \  t=1,2,\ldots,
\end{equation}
\end{shrinkeq}
where $\bm{e}^{\textnormal{G}}_{t}$ denotes the feature maps of the $t^{\textnormal{th}}$ frame. Then a recurrent policy network $\pi$ is learned to aggregate the features of all previous frames $\{\bm{e}^{\textnormal{G}}_{1}, \ldots, \bm{e}^{\textnormal{G}}_{t}\}$, and accordingly determine the location of a small image patch $\tilde{\bm{v}}_t$ to capture the most informative region in $\bm{v}_t$ for a given task.

The selected patch $\tilde{\bm{v}}_t$ will be fed into a high-capacity, accurate but computationally more expensive local encoder $f_{\textnormal{L}}$ to extract the local feature maps $\bm{e}^{\textnormal{L}}_{t}$:
\begin{shrinkeq}{-0ex}
\begin{equation}
    \label{eq:local_encoder}
    \bm{e}^{\textnormal{L}}_{t} = f_{\textnormal{L}}(\tilde{\bm{v}}_t),\ \ \  t=1,2,\ldots.
\end{equation}
\end{shrinkeq}
Importantly, the computational cost introduced by Eq.(\ref{eq:local_encoder}) is considerably smaller than activating $f_{\textnormal{L}}$ for processing the whole frame due to the reduced size of $\tilde{\bm{v}}_t$. AdaFocus is designed to unevenly allocate the computation across the spatial dimension to improve inference efficiency.

Finally, a classifier $f_{\textnormal{C}}$ integrates the information from all the processed frames, and produces the softmax prediction $\bm{p}_t$ at $t^{\textnormal{th}}$ step, i.e.,
\begin{shrinkeq}{-0ex}
\begin{equation}
    \label{eq:classifier}
    \bm{p}_t = f_{\textnormal{C}}(\textnormal{cat}({\bm{e}}^{\textnormal{G}}_{1}, {\bm{e}}^{\textnormal{L}}_{1}),\ldots,\textnormal{cat}({\bm{e}}^{\textnormal{G}}_{t}, {\bm{e}}^{\textnormal{L}}_{t})),
\end{equation}
\end{shrinkeq}
where $\textnormal{cat}(\cdot)$ is the concatenation operation\footnote{Typically, global average pooling will be performed on ${\bm{e}}^{\textnormal{G}}_{t}, {\bm{e}}^{\textnormal{L}}_{t}$ before concatenation to obtain feature vectors. Here we omit this for simplicity.}. Note that $\bm{e}^{\textnormal{G}}_{t}$ is leveraged for both localizing the informative patches and classification, under the goal of facilitating efficient feature reuse. This design is natural since it has been observed that deep networks (e.g., ConvNets and Vision Transformers) excel at learning representations for both recognition and localization simultaneously \cite{zhou2016learning, selvaraju2017grad, dosovitskiy2021an}. In addition, the architecture of $f_{\textnormal{C}}$ may have different choices, such as recurrent networks \cite{hochreiter1997long, cho-etal-2014-learning}, averaging the frame-wise predictions \cite{lin2019tsm, meng2020ar, meng2021adafuse}, and accumulated feature pooling \cite{ghodrati2021frameexit}.

\textbf{Training algorithm.}
In the original AdaFocus \cite{Wang_2021_ICCV} (referred to as AdaFocusV1), the task of selecting task-relevant patches is modeled as a non-differentiable decision problem on several pre-defined patch candidates. Hence, the training of AdaFocusV1 includes both continuous (i.e., video recognition) and discrete (i.e., patch localization) optimization, resulting in a 3-stage algorithm to solve it alternatively. They first train the classification components (i.e., $f_{\textnormal{G}}$, $f_{\textnormal{L}}$ and $f_{\textnormal{C}}$) with random patches, and then fix them to learn a patch selection strategy (i.e., $\pi$) using reinforcement learning. The final stage further fine-tunes the model with the learned policy.

\textbf{Limitations of AdaFocusV1.}
The underlying logic of the 3-stage training is straightforward. However, this procedure is unfriendly for practitioners. First, effectively deploying the reinforcement learning algorithm is nontrivial. It requires considerable efforts for properly designing the key components (e.g., the action space and the reward function), and implementing specialized optimization techniques (e.g., deep Q-Network \cite{mnih2013playing} or proximal policy optimization \cite{schulman2017proximal}). Second, the 3-stage alternative algorithm is an indirect formulation for optimizing the recognition objective, which tends to be time-consuming, and may result in sub-optimal solutions. Third, the performance of AdaFocusV1 largely depends on a number of implementation configurations (e.g., performing pre-training, freezing some components in different stages, and stage-wise hyper-parameter searching) that need to be carefully tuned on a per-dataset or per-backbone basis.

In the following, we present an end-to-end trainable formulation for AdaFocus to address the issue of inefficient training. The proposed network, AdaFocusV2, can be conveniently implemented to achieve consistently better performance than AdaFocusV1 with reduced training cost.

\begin{figure}[t]
    \begin{center}
    \centerline{\includegraphics[width=0.96\columnwidth]{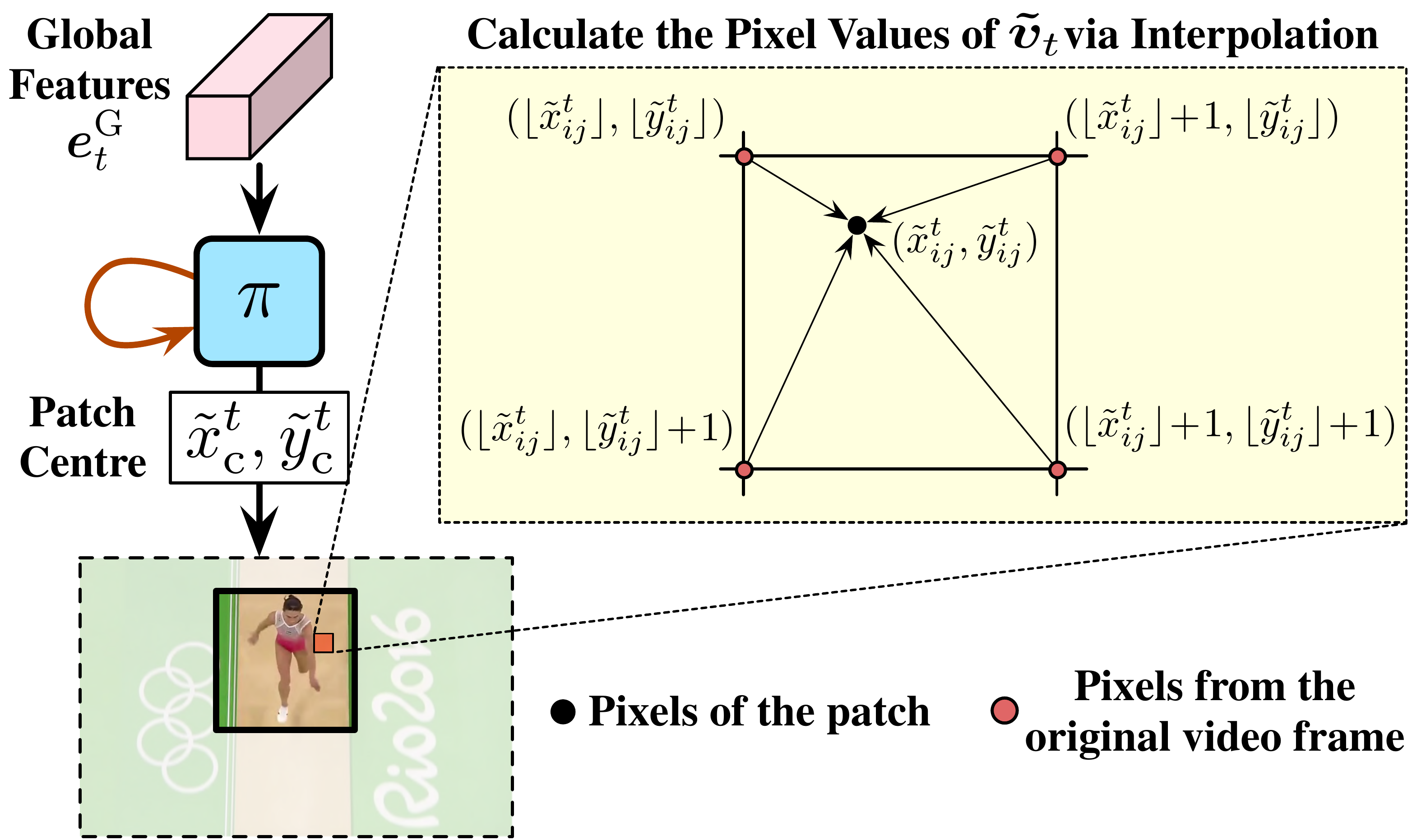}}
    \caption{\textbf{Illustration of interpolation-based patch selection.} This operation is differentiable, i.e., the gradients can be directly back-propagated into the policy network $\pi$ through the selected image patch $\tilde{\bm{v}}_t$. Consequently, integrating the learning of $\pi$ into a unified end-to-end training paradigm turns out to be feasible. \label{fig:interpolation}
    }
    \end{center}
    \vskip -0.25in
\end{figure}

\subsection{Interpolation-based Patch Selection}
\label{sec:32}


To enable end-to-end training, we propose a differentiable solution to obtain $\tilde{\bm{v}}_t$. Suppose that the size of the original frame ${\bm{v}}_t$ and the patch $\tilde{\bm{v}}_t$ is $H\!\times\!W$ and $P\!\times\!P\ (P\!<\!H, W)$, respectively\footnote{In our implementation, the height/width/coordinates are correspondingly normalized using the linear projection $[0,H]\!\to\![0,1]$ and $[0,W]\!\to\![0,1]$. Here we use the original values for the ease of understanding.}. We assume that $\pi$ outputs the continuous centre coordinates $(\tilde{x}^t_{\textnormal{c}}, \tilde{y}^t_{\textnormal{c}})$ of $\tilde{\bm{v}}_t$, namely
\begin{shrinkeq}{-0ex}
\begin{equation}
    \label{eq:cetre_xy}
    \begin{split}
        &(\tilde{x}^t_{\textnormal{c}},\ \tilde{y}^t_{\textnormal{c}}) = \pi({\bm{e}}^{\textnormal{G}}_{1},\ldots, {\bm{e}}^{\textnormal{G}}_{t}), \\
        \tilde{x}^t_{\textnormal{c}} \in [&\frac{P}{2}, W-\frac{P}{2}],\ \ \ \tilde{y}^t_{\textnormal{c}} \in [\frac{P}{2}, H-\frac{P}{2}],
    \end{split}
\end{equation}
\end{shrinkeq}
where ${\bm{e}}^{\textnormal{G}}_{1},\ldots, {\bm{e}}^{\textnormal{G}}_{t}$ are the global features of $1^{\textnormal{st}}-t^{\textnormal{th}}$ frames extracted by the global encoder $f_{\textnormal{G}}$. Notably, we refer to the coordinates of the top-left corner of the frame as $(0,0)$, and Eq. (\ref{eq:cetre_xy}) ensures that $\tilde{\bm{v}}_t$ will never go outside of ${\bm{v}}_t$. Our aim is to calculate the values of all pixels in $\tilde{\bm{v}}_t$, while allowing the gradients to be back-propagated through $(\tilde{x}^t_{\textnormal{c}}, \tilde{y}^t_{\textnormal{c}})$. 

\textbf{Feed-forward.}
We first introduce the feed-forward process of our method. Formally, the coordinates of a pixel in the patch $\tilde{\bm{v}}_t$ can be expressed as the addition of $(\tilde{x}^t_{\textnormal{c}}, \tilde{y}^t_{\textnormal{c}})$ and a fixed offset:
\begin{shrinkeq}{-0ex}
\begin{equation}
    \label{eq:c_plus_offset}
    \begin{split}
        &(\tilde{x}^t_{ij},\ \tilde{y}^t_{ij}) = (\tilde{x}^t_{\textnormal{c}},\ \tilde{y}^t_{\textnormal{c}}) + \bm{o}_{ij}, \\
        \bm{o}_{ij} &\in {\left\{ -\frac{P}{2}, -\frac{P}{2} + 1, \ldots, \frac{P}{2} \right\}}^2.
    \end{split}
\end{equation}
\end{shrinkeq}
Herein, $(\tilde{x}^t_{ij}, \tilde{y}^t_{ij})$ denotes the horizontal and vertical coordinates in the original frame ${\bm{v}}_t$ corresponding to the pixel in the $i^{\textnormal{th}}$ row and $j^{\textnormal{th}}$ column of $\tilde{\bm{v}}_t$, while $\bm{o}_{ij}$ represents the vector from the patch centre $(\tilde{x}^t_{\textnormal{c}}, \tilde{y}^t_{\textnormal{c}})$ to this pixel. Given a fixed patch size, $\bm{o}_{ij}$ is a constant conditioned only on $i,j$, regardless of $t$ or the inputs of $\pi$.

Since the values of $(\tilde{x}^t_{\textnormal{c}}, \tilde{y}^t_{\textnormal{c}})$ are continuous, there does not exist a pixel of ${\bm{v}}_t$ exactly located at $(\tilde{x}^t_{ij}, \tilde{y}^t_{ij})$ to directly get the pixel value. Alternatively, as illustrated in Figure \ref{fig:interpolation}, we can always find that the location $(\tilde{x}^t_{ij}, \tilde{y}^t_{ij})$ is surrounded by four adjacent pixels of ${\bm{v}}_t$, forming a grid. The coordinates are $(\lfloor\tilde{x}^t_{ij}\rfloor, \lfloor\tilde{y}^t_{ij}\rfloor)$,  $(\lfloor\tilde{x}^t_{ij}\rfloor\!+\!1, \lfloor\tilde{y}^t_{ij}\rfloor)$,  $(\lfloor\tilde{x}^t_{ij}\rfloor, \lfloor\tilde{y}^t_{ij}\rfloor\!+\!1)$ and $(\lfloor\tilde{x}^t_{ij}\rfloor\!+\!1, \lfloor\tilde{y}^t_{ij}\rfloor\!+\!1)$, respectively, where $\lfloor\cdot\rfloor$ denotes the rounding-down operation. By assuming that the corresponding pixel values of these four pixels are $(m^t_{ij})_{00}$, $(m^t_{ij})_{01}$, $(m^t_{ij})_{10}$, and $(m^t_{ij})_{11}$, the pixel value at $(\tilde{x}^t_{ij}, \tilde{y}^t_{ij})$ (referred to as $\tilde{m}^t_{ij}$) can be obtained via interpolation algorithms. In this paper, we simply adopt the differentiable bilinear interpolation:
\begin{shrinkeq}{-0ex}
\begin{equation}
    \label{eq:bilinear}
    \begin{split}
        \tilde{m}^t_{ij} &= (m^t_{ij})_{00}(\lfloor\tilde{x}^t_{ij}\rfloor\!-\!\tilde{x}^t_{ij}\!+\!1)(\lfloor\tilde{y}^t_{ij}\rfloor\!-\!\tilde{y}^t_{ij}\!+\!1) \\
        &+(m^t_{ij})_{01}(\tilde{x}^t_{ij}\!-\!\lfloor\tilde{x}^t_{ij}\rfloor)(\lfloor\tilde{y}^t_{ij}\rfloor\!-\!\tilde{y}^t_{ij}\!+\!1) \\
        &+(m^t_{ij})_{10}(\lfloor\tilde{x}^t_{ij}\rfloor\!-\!\tilde{x}^t_{ij}\!+\!1)(\tilde{y}^t_{ij}\!-\!\lfloor\tilde{y}^t_{ij}\rfloor) \\
        &+(m^t_{ij})_{11}(\tilde{x}^t_{ij}\!-\!\lfloor\tilde{x}^t_{ij}\rfloor)(\tilde{y}^t_{ij}\!-\!\lfloor\tilde{y}^t_{ij}\rfloor).
    \end{split}
\end{equation}
\end{shrinkeq}
Consequently, we can obtain the image patch $\tilde{\bm{v}}_t$ by traversing all possible $i,j$ in Eq. (\ref{eq:bilinear}).

\textbf{Back-propagation.}
Give the training loss $\mathcal{L}$, it is easy to compute the gradient ${\partial\mathcal{L}}/{\partial\tilde{m}^t_{ij}}$ with standard back-propagation. Then, following on the chain rule, we have 
\begin{shrinkeq}{-0ex}
\begin{equation}
        \label{eq:bp_1}
        \frac{\partial\mathcal{L}}{\partial\tilde{x}^t_{\textnormal{c}}} \!=\!\! \sum_{i,j}\! \frac{\partial\mathcal{L}}{\partial\tilde{m}^t_{ij}}
        \frac{\partial\tilde{m}^t_{ij}}{\partial\tilde{x}^t_{\textnormal{c}}}, \ \ \ 
        \frac{\partial\mathcal{L}}{\partial\tilde{y}^t_{\textnormal{c}}} \!=\!\! \sum_{i,j}\! \frac{\partial\mathcal{L}}{\partial\tilde{m}^t_{ij}}
        \frac{\partial\tilde{m}^t_{ij}}{\partial\tilde{y}^t_{\textnormal{c}}}.
\end{equation}
\end{shrinkeq}
Combining Eq. (\ref{eq:c_plus_offset}) and Eq. (\ref{eq:bp_1}), we can further derive
\begin{shrinkeq}{-0ex}
\begin{equation}
    \label{eq:bp_2}
    \frac{\partial\tilde{m}^t_{ij}}{\partial\tilde{x}^t_{\textnormal{c}}}\!=\!\frac{\partial\tilde{m}^t_{ij}}{\partial\tilde{x}^t_{ij}},\ \ \ 
    \frac{\partial\tilde{m}^t_{ij}}{\partial\tilde{y}^t_{\textnormal{c}}}\!=\!\frac{\partial\tilde{m}^t_{ij}}{\partial\tilde{y}^t_{ij}}.
\end{equation}
\end{shrinkeq}
Eq. (\ref{eq:bp_2}) can be solved by leveraging Eq. (\ref{eq:bilinear}), such that we can obtain the gradients ${\partial\mathcal{L}}/{\partial\tilde{x}^t_{\textnormal{c}}}$ and ${\partial\mathcal{L}}/{\partial\tilde{y}^t_{\textnormal{c}}}$. Given that $\tilde{x}^t_{\textnormal{c}}$ and $\tilde{y}^t_{\textnormal{c}}$ are the outputs of the policy network $\pi$, the back-propagation process is able to proceed in an ordinary way.

\subsection{Training Techniques}

\textbf{Naive implementation.}
Thus far, we have enabled the gradients to be back-propagated throughout the whole AdaFocus network for updating all trainable parameters simultaneously. Consequently, end-to-end training has been feasible. For example, one can minimize the frame-wise cross-entropy loss $L_{\textnormal{CE}}(\cdot)$ in AdaFocusV1 \cite{Wang_2021_ICCV}:
\begin{shrinkeq}{-0ex}
\begin{equation}
    \label{eq:naive_objective}
        \mathop{\textnormal{minimize}}_{f_{\textnormal{G}}, f_{\textnormal{L}}, f_{\textnormal{C}}, \pi}\ \ \mathcal{L}=\mathop{\mathbb{E}}_{\{\bm{v}_1, \bm{v}_2, \ldots\}}
    [
        \frac{1}{T}\sum\nolimits_{t=1}^{T} L_{\textnormal{CE}}(\bm{p}_t, y)
    ], 
\end{equation}
\end{shrinkeq}
where $T$ and $y$ denote the length and the label of the video $\{\bm{v}_1, \bm{v}_2, \ldots\}$, and $\bm{p}_t$ is the softmax prediction at $t^{\textnormal{th}}$ frame, as stated in Section \ref{sec:31}.

However, importantly, such a straightforward implementation leads to the severely degraded performance (see Table \ref{tab:training_tech} for experimental evidence). We attribute this issue to the absence of some appealing optimization properties introduced by 3-stage training, namely the lack of \emph{supervision}, \emph{input diversity} and \emph{training stability}. To solve these problems, we develop three simple but effective training techniques, with which end-to-end training can significantly outperform the 3-stage counterpart. These techniques do not introduce additional tunable hyper-parameters, while achieving consistent improvements across varying datasets, backbone architectures, and model configurations.

\textbf{\emph{Lack of supervision}: auxiliary supervision.}
AdaFocusV1 first pre-trains the two encoders (i.e., $f_{\textnormal{G}}, f_{\textnormal{L}}$) separately using a direct frame-wise recognition loss (by simply appending a fully-connected layer), aiming to obtain a proper initialization. In contrast, when solving problem (\ref{eq:naive_objective}), $f_{\textnormal{G}}$ and $f_{\textnormal{L}}$ are trained from scratch, while they are only indirectly supervised by the gradients from the classifier $f_{\textnormal{C}}$. We find that explicitly introducing auxiliary supervision on $f_{\textnormal{G}}$ and $f_{\textnormal{L}}$ effectively facilitates the efficient end-to-end training of AdaFocusV2. In specific, we attach two linear classifiers, $\textnormal{FC}_{\textnormal{G}}(\cdot)$ and $\textnormal{FC}_{\textnormal{L}}(\cdot)$, to the outputs of $f_{\textnormal{G}}$ and $f_{\textnormal{L}}$, and replace the loss function $\mathcal{L}$ in (\ref{eq:naive_objective}) by $\mathcal{L}'$:
\begin{shrinkeq}{-0ex}
\begin{equation}
    \label{eq:l_prime}
    \begin{split}
    \mathcal{L}'=\mathop{\mathbb{E}}_{\{\bm{v}_1, \bm{v}_2, \ldots\}}
    \{
        &\frac{1}{T}\sum\nolimits_{t=1}^{T} 
        \left[
            L_{\textnormal{CE}}(\bm{p}_t, y)\right. \\ 
            &+ L_{\textnormal{CE}}(\textnormal{SoftMax}(\textnormal{FC}_{\textnormal{G}}(\overline{\bm{e}}^{\textnormal{G}}_{t})), y) \\
            &+ \left.L_{\textnormal{CE}}(\textnormal{SoftMax}(\textnormal{FC}_{\textnormal{L}}(\overline{\bm{e}}^{\textnormal{L}}_{t})), y) 
        \right]
    \}, 
    \end{split}
\end{equation}
\end{shrinkeq}
where $\overline{\bm{e}}^{\textnormal{G}}_{t}$ and $\overline{\bm{e}}^{\textnormal{L}}_{t}$ are the feature vectors after performing global average pooling on the feature maps ${\bm{e}}^{\textnormal{G}}_{t}$ and ${\bm{e}}^{\textnormal{L}}_{t}$ output by $f_{\textnormal{G}}$ and $f_{\textnormal{L}}$. Intuitively,
minimizing $\mathcal{L}'$ enforces the two encoders to learn linearized deep representations, which has been widely verified as an efficient approach for training deep networks \cite{he2016deep, huang2019convolutional, dosovitskiy2021an}. This paradigm benefits the learning of $f_{\textnormal{C}}$ as well, since its inputs are explicitly regularized to be linearly separable.

\textbf{\emph{Lack of input diversity}: diversity augmentation.}
In the stage-1 for training AdaFocusV1, image patches are randomly cropped, yielding highly diversified inputs for learning well-generalized local encoder $f_{\textnormal{L}}$. However, the patch selection process presented in Section \ref{sec:32} is deterministic. In Eq. (\ref{eq:l_prime}), given a video frame, the local encoder $f_{\textnormal{L}}$ only has access to the patch specified by the policy network $\pi$. This procedure leads to the limited diversity of training data for the inputs of $f_{\textnormal{L}}$. Empirically, we observe that it results in the inferior performance of $f_{\textnormal{L}}$. We address this issue by proposing a straightforward diversity augmentation approach. For each video, we first compute $\mathcal{L}'$ by activating $\pi$ as aforementioned. Then we infer $f_{\textnormal{L}}$ and the classifier $f_{\textnormal{C}}$ for a second time using randomly cropped patches, obtaining an additional loss $\mathcal{L}'_{\textnormal{random}}$, which follows Eq. (\ref{eq:l_prime}) as well. Our final optimization objective is to minimize the combination of $\mathcal{L}'$ and $\mathcal{L}'_{\textnormal{random}}$:
\begin{shrinkeq}{-0ex}
\begin{equation}
    \label{eq:final_objective}
        \mathop{\textnormal{minimize}}_{f_{\textnormal{G}}, f_{\textnormal{L}}, f_{\textnormal{C}}, \pi}\ \ 
        \frac{1}{2} (\mathcal{L}'+\mathcal{L}'_{\textnormal{random}}).
\end{equation}
\end{shrinkeq}

\textbf{\emph{Lack of training stability}: stop-gradient.}
In AdaFocusV1, the policy network $\pi$ is learned on top of the fixed and completely trained global encoder $f_{\textnormal{G}}$. When it comes to end-to-end training, $\pi$ and $f_{\textnormal{G}}$ are simultaneously updated. In this case, we observe that the gradients back-propagated from $\pi$ interfere with the learning of $f_{\textnormal{G}}$, leading to an unstable training process with slow convergence speed. We find that this problem can be solved by simply stopping the gradients before the inputs of $\pi$. In other words, we propose to train $f_{\textnormal{G}}$ using the pure classification objective without any effect from $\pi$, as done in AdaFocusV1. This design is rational since previous works have revealed that the representations extracted by deep recognition networks can naturally be leveraged for localizing task-relevant regions \cite{zhou2016learning, selvaraju2017grad, dosovitskiy2021an}. 


\begin{figure}[t]
    \begin{center}
    \centerline{\includegraphics[width=\columnwidth]{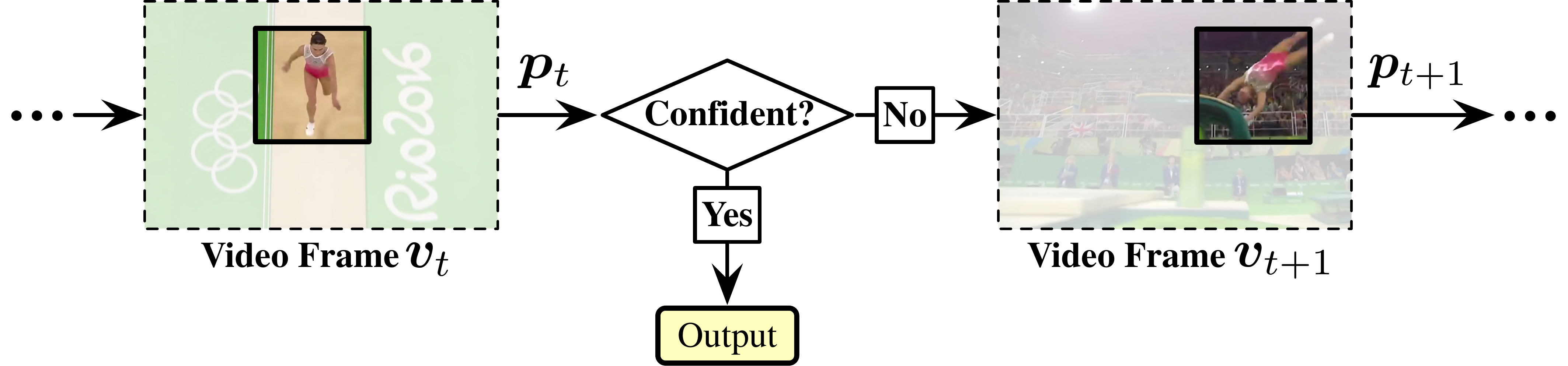}}
    \vskip -0.025in
    \caption{\textbf{Illustration of AdaFocusV2+.}  \label{fig:AdaFocusV2_plus}
    }
    \end{center}
    \vskip -0.42in
\end{figure}

\begin{figure*}[!t]
    \begin{center}
    \begin{minipage}{0.959\columnwidth}
    \centering
    \begin{footnotesize}
    \captionof{table}{\textbf{Comparisons of AdaFocusV1 and AdaFocusV2 on training efficiency.} ActivityNet mAP and wall-clock training time are reported. The latter is obtained using 4 NVIDIA 3090 GPUs. The better results are \textbf{bold-faced} (E2E: End-to-End).}
    \label{tab:actnet_vs_v1_train}
    \vskip -0.18in
    \setlength{\tabcolsep}{1.2mm}{
    \vspace{5pt}
    \renewcommand\arraystretch{0.925}
    \resizebox{\columnwidth}{!}{
    \begin{tabular}{c|c|ccc|ccc}
    \toprule
    \multirow{2}{*}{Methods} &  3-stage  & \multicolumn{3}{c|}{mAP on ActivityNet} &  \multicolumn{3}{|c}{Training Wall-time} \\
    & Training & 96$^2$ & 128$^2$ & 160$^2$ & 96$^2$ & 128$^2$ & 160$^2$ \\
    \midrule
    AdaFocusV1-MN2/RN & \textcolor{red}{\cmark} & \ 71.9\% & 75.0\% & 76.0\% &\  6.4h & 7.2h & 8.6h \\
    AdaFocusV2-MN2/RN & \textcolor{green}{\xmark} & \ \textbf{77.4\%} & \textbf{79.0\%} & \textbf{79.9\%} &\  \textbf{3.4h} & \textbf{3.7h} & \textbf{4.3h} \\
    \bottomrule
    \end{tabular}}}
    \end{footnotesize}
        \end{minipage}    
    \hspace{0.005in}
    \begin{minipage}{1.115\columnwidth}
        \vskip -0.02in
        \centering
        \begin{footnotesize}
        \captionof{table}{\textbf{AdaFocusV1+ v.s. AdaFocusV2+.} The two methods are deployed on the same AdaFocusV1 base model to reduce temporal redundancy. ActivityNet mAP on varying inference cost (GFLOPs/video) is reported. The better results are \textbf{bold-faced} (RL: reinforcement learning).}
        \label{tab:actnet_vs_v1_plus}
        \vskip -0.175in
        \setlength{\tabcolsep}{1.225mm}{
        \vspace{5pt}
        \renewcommand\arraystretch{0.925}
        \resizebox{0.99\columnwidth}{!}{
        \begin{tabular}{c|c|cc|cc}
        \toprule
        \multirow{2}{*}{Methods} & Additional  & \multicolumn{2}{c|}{On AdaFocusV1 (128$^2$)} &  \multicolumn{2}{|c}{On AdaFocusV1 (160$^2$)} \\
        & Training&  19.6G/video & 24.4G/video &  27.2G/video & 34.6G/video \\
        \midrule
        AdaFocusV1-MN2/RN+ & \textcolor{red}{\cmark \emph{(RL)}} &  74.2\% & 74.8\% &  75.5\% & 75.9\% \\
        AdaFocusV2-MN2/RN+ & \textcolor{green}{\xmark} &  \textbf{74.6\%} & \textbf{75.0\%} &  {75.5\%} & {75.9\%} \\
        \bottomrule
        \end{tabular}}}
        \end{footnotesize}     
    \end{minipage}
    \end{center}
    \vskip -0.2in
 \end{figure*}

 \begin{table*}[t]
    \centering
    \vskip -0.03in
    \begin{footnotesize}
    \caption{\textbf{Comparisons of AdaFocusV2 and state-of-the-art baselines on ActivityNet, FCVID and Mini-Kinetics.} GFLOPs refers to the average computational cost for processing a single video. The best two results are \textbf{bold-faced} and \underline{underlined}, respectively.}
    \label{tab:actnet_main_table}
    \vskip -0.18in
    \setlength{\tabcolsep}{2mm}{
    \vspace{5pt}
    \renewcommand\arraystretch{0.9}
    \resizebox{1.85\columnwidth}{!}{
    \begin{tabular}{c|c|c|cccccc}
    \toprule
    \multirow{2}{*}{Methods} & \multirow{2}{*}{Published on}& \multirow{2}{*}{Backbones}  & \multicolumn{2}{c}{ActivityNet} &  \multicolumn{2}{c}{FCVID} & \multicolumn{2}{c}{Mini-Kinetics}  \\
    &&& mAP & GFLOPs & mAP &  GFLOPs & Top-1 Acc. &  GFLOPs \\
    \midrule
    AdaFuse \cite{meng2021adafuse} & \emph{ICLR'21} & ResNet & \ \ 73.1\% &  61.4 & \ \ \underline{81.6\%} & 45.0 & \ \ 72.3\% &  23.0 \\
    Dynamic-STE \cite{kim2021efficient} & \emph{ICCV'21} & ResNet  & \ \ 75.9\% & 30.5 & -- & -- & \ \  72.7\% & \underline{18.3} \\
    FrameExit \cite{ghodrati2021frameexit} & \emph{CVPR'21} & ResNet  & \ \ \underline{76.1\%} & \underline{26.1}  & -- & -- & \ \ {72.8\%} & {19.7}  \\
    \midrule
    AdaFocusV2-RN (128$^2$)  & -- & ResNet  & \ \ \textbf{78.9\%} & 34.1 & \ \ \textbf{84.5\%}& \underline{34.1} &\ \ \textbf{74.0\%}& {34.1}  \\
    AdaFocusV2-RN+ (128$^2$)  & -- & ResNet  & \ \ \underline{76.1\%} & \textbf{15.3} &\ \ \underline{81.6\%}& \textbf{12.0}& \ \ \underline{72.8\%}& \textbf{13.4}\\
    \midrule
    \midrule
    LiteEval \cite{wu2019liteeval} & \emph{NeurIPS'19} & MobileNet-V2 + ResNet  & \ \ 72.7\% & 95.1 & \ \ 80.0\% & 94.3  & \ \ 61.0\% & 99.0 \\
    SCSampler \cite{korbar2019scsampler} & \emph{ICCV'19} & MobileNet-V2 + ResNet & \ \ 72.9\%  & 42.0 & \ \ 81.0\% & 42.0  & \ \ 70.8\%  & 42.0 \\
    ListenToLook \cite{gao2020listen} & \emph{CVPR'20} & MobileNet-V2 + ResNet & \ \ 72.3\%  & 81.4 & -- & --  & -- & --  \\
    AR-Net \cite{meng2020ar} & \emph{ECCV'20} & MobileNet-V2 + ResNet & \ \ 73.8\% & 33.5 & \ \ 81.3\% & 35.1  & \ \ 71.7\% & 32.0  \\
    AdaFrame \cite{wu2019adaframe} & \emph{T-PAMI'21} & MobileNet-V2 + ResNet & \ \ 71.5\% & 79.0 & \ \ 80.2\% & 75.1  & -- & --  \\
    VideoIQ \cite{sun2021dynamic} & \emph{ICCV'21} & MobileNet-V2 + ResNet & \ \ \underline{74.8\%} &  28.1 & \ \ \underline{82.7\%} & \underline{27.0}  & \ \ \underline{72.3\%} &  \underline{20.4} \\
    \midrule
    AdaFocusV2-MN2/RN (128$^2$)  & -- & MobileNet-V2 + ResNet  & \ \ \textbf{79.0\%} & \underline{27.0} &\ \ \textbf{85.0\%}& \underline{27.0} &\ \ \textbf{75.4\%}& 27.0  \\
    AdaFocusV2-MN2/RN+ (128$^2$)  & -- & MobileNet-V2 + ResNet  &\ \ \underline{74.8\%} & \textbf{9.9} &\ \ \underline{82.7\%}&\textbf{10.1}&\ \ \underline{72.3\%}& \textbf{6.3} \\
    \bottomrule
    \end{tabular}}}
    \end{footnotesize}
    \vskip -0.13in
\end{table*}


\subsection{Reducing Temporal Redundancy}
\label{sec:v2_plus}



The basic formulation of AdaFocus processes each frame using the same amount of computation, and hence it can be improved by further reducing temporal redundancy. AdaFocusV1 achieves this via skipping less informative frames with reinforcement learning. In contrast, we propose a simple confidence-based early-exit algorithm that achieves competitive performance. Our approach can be directly deployed on AdaFocusV2 trained following the aforementioned paradigm, without any additional training process. We refer to this extension of AdaFocusV2 as AdaFocusV2+, as shown in Figure \ref{fig:AdaFocusV2_plus}. 

It has been widely observed that there exist a considerable number of ``easier'' samples in datasets \cite{huang2017multi, yang2020resolution, NeurIPS2020_7866, ghodrati2021frameexit, wang2021not, li2021ds}, which can be accurately recognized with smaller computational cost than others. In the context of videos, we assume that processing a subset of frames (from the beginning) rather than all may be sufficient for these ``easier'' samples. To implement this idea, at test time, we propose to compare the largest entry of the softmax prediction $\bm{p}_t$ ({defined as confidence in previous works \cite{huang2017multi, yang2020resolution, NeurIPS2020_7866, wang2021not}}) at $t^{\textnormal{th}}$ frame with a pre-defined threshold $\eta_t$. Once $\max_j p_{tj} \geq \eta_t$, the prediction will be postulated to be reliable enough, and the inference will be terminated by outputting $\bm{p}_t$. We always adopt a zero-threshold for at final frame.

The values of $\{\eta_1, \eta_2, \ldots\}$ are solved on the validation set. Suppose that the model needs to classify a set of samples $\mathcal{D}_{\textnormal{val}}$ within a given computational budget $B>0$ \cite{huang2017multi, wang2021not}. One can obtain the thresholds through
\begin{shrinkeq}{-0ex}
\begin{equation}
    \label{eq:thres}
    \begin{split}
    \mathop{\textnormal{maximize}}_{\eta_1, \eta_2, \ldots}&\ \ \ \textnormal{Acc}(\eta_1, \eta_2, \ldots|\mathcal{D}_{\textnormal{val}})\\ 
    \textnormal{subject to}&\ \ \ \textnormal{FLOPs}(\eta_1, \eta_2, \ldots|\mathcal{D}_{\textnormal{val}})\leq B.
    \end{split}
\end{equation}
\end{shrinkeq}

Here $\textnormal{Acc}(\eta_1, \eta_2, \ldots|\mathcal{D}_{\textnormal{val}})$ and $\textnormal{FLOPs}(\eta_1, \eta_2, \ldots|\mathcal{D}_{\textnormal{val}})$ refer to the accuracy and computational cost on $\mathcal{D}_{\textnormal{val}}$ using the thresholds $\{\eta_1, \eta_2, \ldots\}$. Notably, by changing $B$, one can obtain varying values of $\{\eta_1, \eta_2, \ldots\}$. The computational cost of AdaFocusV2+ can be flexibly adjusted without additional training by simply adjusting these thresholds. In our implementation, we solve problem (\ref{eq:thres}) following the method proposed in \cite{huang2017multi} on training set, which we find performs on par with using cross-validation.

\begin{figure*}[!t]
    \begin{center}
    \begin{minipage}{1.37\columnwidth}
        \hspace{-0.05in}
        \includegraphics[width=1.015\columnwidth]{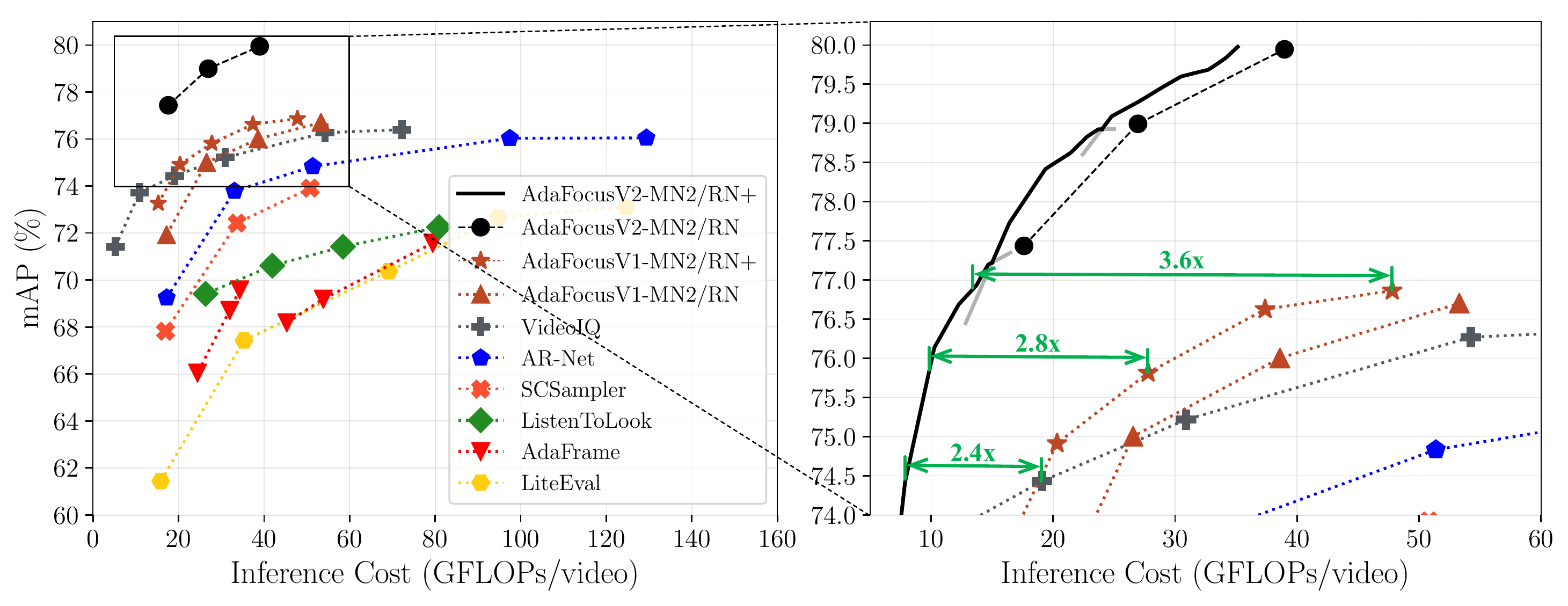}	
        \vskip -0.11in
        \caption{\textbf{Comparisons of AdaFocusV2, AdaFocusV1 and state-of-the-art baselines on ActivityNet in terms of inference efficiency.} MobileNetV2 (MN2) and ResNet (RN) are deployed as backbones in all methods. AdaFocusV2 adopts 96$^2$, 128$^2$, and 160$^2$ patches. Notably, AdaFocusV2+ can switch within each black curve without additional training.
    }\label{fig:actnet_vs_sota}
    \end{minipage} 
    \hspace{0.01in}
    \begin{minipage}{0.7\columnwidth}
        \vskip 0.01in
        \hspace{-0.1in}
        \includegraphics[width=1.035\columnwidth]{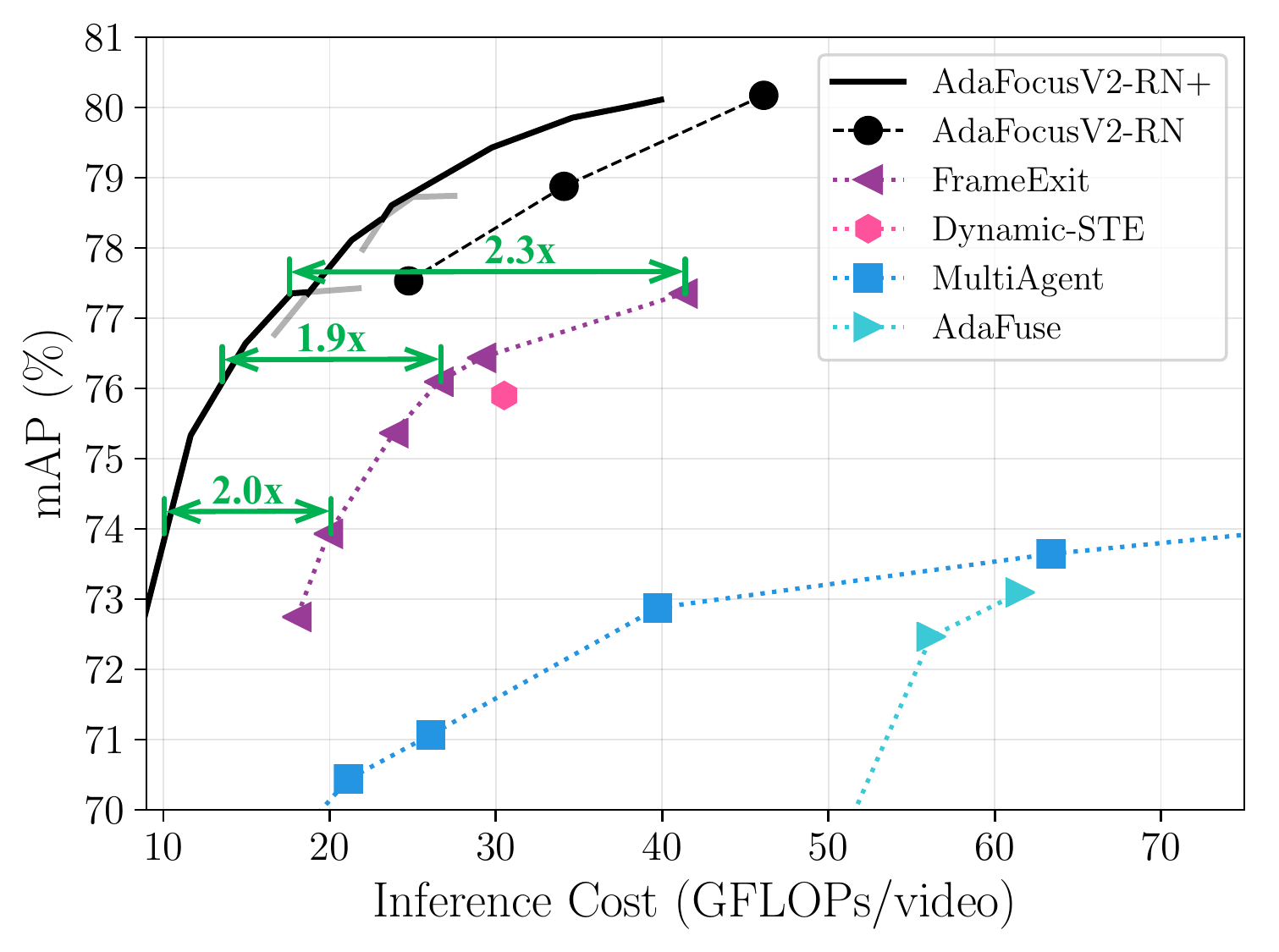}
        \vskip -0.1in
        \caption{\textbf{AdaFocusV2 (96$^2$, 128$^2$, 160$^2$) v.s. state-of-the-art efficient video recognition frameworks on top of ResNet (RN).} Results on ActivityNet are reported.
    }\label{fig:actnet_vs_v1}
    \end{minipage}   
    \end{center}
    \vskip -0.25in
 \end{figure*}

%% file: experiment.tex
\vspace{-0.5ex}
\section{Experiment}
\vspace{-0.5ex}


In this section, we empirically compare AdaFocusV2 with both AdaFocusV1 and state-of-the-art efficient video recognition frameworks. We also demonstrate that AdaFocusV2 is able to effectively reduce the inference cost on top of recent representative light-weighted deep networks. In addition, ablation and visualization results are provided for more insights. Code and pre-trained models are available at \url{https://github.com/LeapLabTHU/AdaFocusV2}.

\textbf{Datasets.}
Six large-scale video recognition datasets are used, i.e., ActivityNet \cite{caba2015activitynet}, FCVID \cite{TPAMI-fcvid}, Mini-Kinetics \cite{kay2017kinetics, meng2020ar}, Something-Something (Sth-Sth) V1\&V2 \cite{goyal2017something} and Jester \cite{materzynska2019jester}. The official training-validation split is adopted. Given the limited space, we introduce them in Appendix A. Following the common practice \cite{wu2019adaframe, lin2019tsm, wu2019liteeval, gao2020listen, meng2020ar, meng2021adafuse, Wang_2021_ICCV}, we evaluate the performance of different methods via mean average precision (mAP) and Top-1 accuracy (Top-1 Acc.) on ActivityNet/FCVID and other datasets, respectively.



\textbf{Setups.}
Unless otherwise specified, we uniformly sample 16 frames from each video on ActivityNet, FCVID and Mini-Kinetics, while sampling 8/12 frames on Sth-Sth. Offline recognition is considered, where we obtain a single prediction for each video. We adopt the final prediction $\bm{p}_{T}$ after processing all frames for AdaFocusV2, and use the early-exited prediction for AdaFocusV2+. We follow the data pre-processing pipeline in \cite{lin2019tsm, meng2020ar,Wang_2021_ICCV}. For inference, we resize all frames to 256$^2$ and perform 224$^2$ centre-crop.



\begin{figure*}[t]
    \begin{center}
        \begin{minipage}{0.7\columnwidth}
            \vskip 0.01in
            \hspace{-0.1in}
            \includegraphics[width=1.0165\columnwidth]{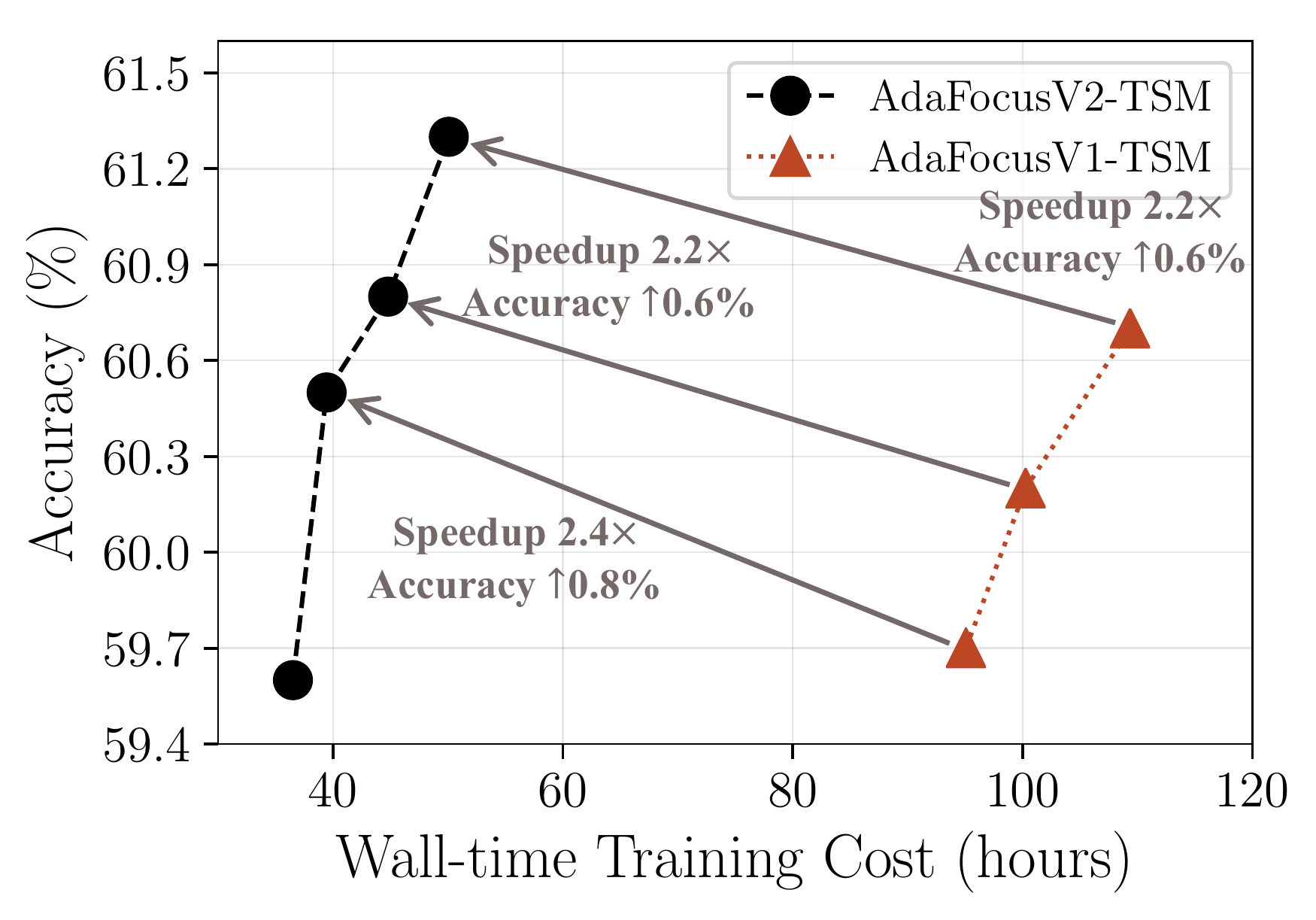}
            \vskip -0.1in
            \caption{\textbf{Comparisons of training efficiency of AdaFocusV1 and AdaFocusV2 on Sth-Sth V2.} The wall-clock training time is obtained based on 4 NVIDIA 3090 GPUs.
        }\label{fig:sthv1_time_acc}
        \end{minipage}    
    \hspace{0.01in}
    \begin{minipage}{1.37\columnwidth}
        \hspace{-0.05in}
        \includegraphics[width=1.015\columnwidth]{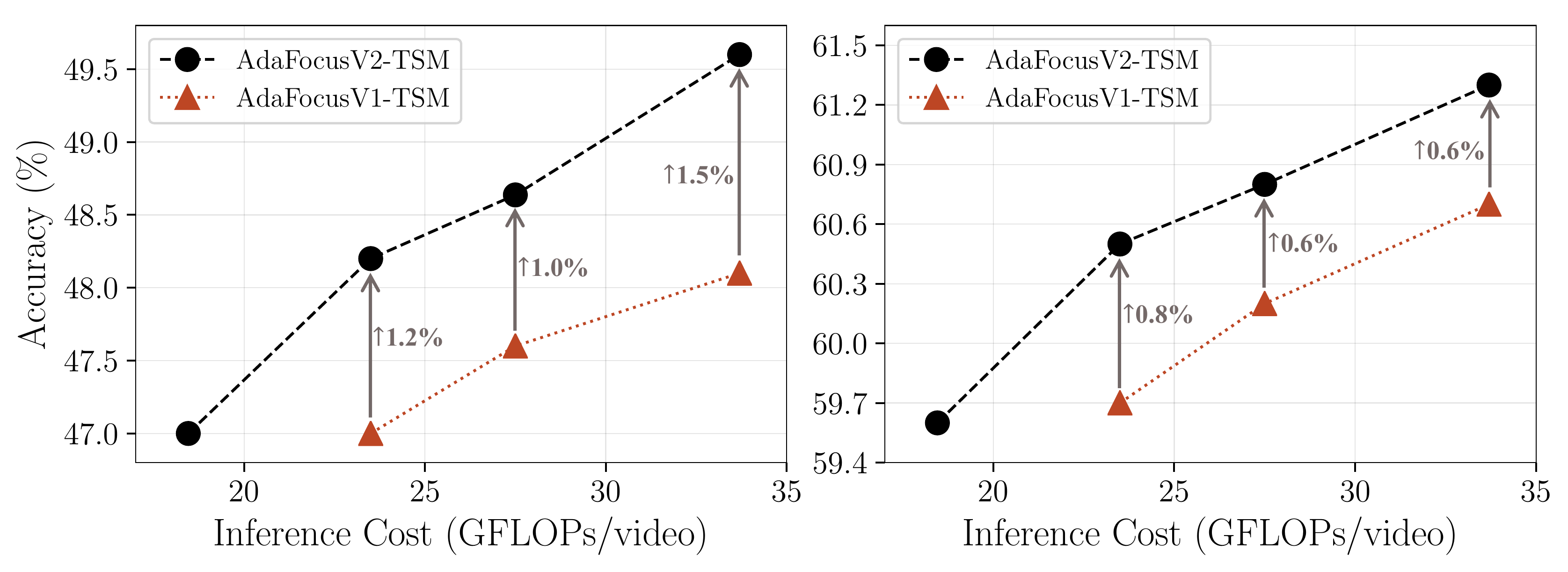}	
        \vskip -0.11in
        \caption{\textbf{AdaFocusV1 v.s. AdaFocusV2 on Sth-Sth V1 (\emph{left}) and Sth-Sth V2 (\emph{right}) in terms of inference efficiency.} The two algorithms are implemented on top of TSM \cite{lin2019tsm} with the patch size of \{144$^2$, 160$^2$, 176$^2$\} and \{128$^2$, 144$^2$, 160$^2$, 176$^2$\}, respectively. AdaFocusV2 outperforms AdaFocusV1 significantly with the same network architecture.
    }\label{fig:sth_flops_acc}
    \end{minipage}
    \end{center}
    \vskip -0.3in
 \end{figure*}

 \begin{table*}[t]
    \centering
    \begin{footnotesize}
    \caption{\textbf{Performance of AdaFocusV2-TSM and representative efficient video recognition models}. MN2, R18/R34/R50 and BN-Inc. denote MobileNet-V2, ResNet-18/34/50 and BN-Inception, respectively. TSM+ refers to the augmented TSM baseline with the same network architecture as our method except for the policy network $\pi$. We uniformly sample 8/12 frames for the MobileNetV2/ResNet-50 in our models. The throughput is tested on an NVIDIA GeForce RTX 3090 GPU with a batch size of 128. The best results are \textbf{bold-faced}.}
    \label{tab:sthsth}
    \vskip -0.2in
    \setlength{\tabcolsep}{0mm}{
    \vspace{5pt}
    \renewcommand\arraystretch{0.9}
    \resizebox{2.02\columnwidth}{!}{
    \begin{tabular}{cccccccccc} 
    \toprule
    \multirow{2}{*}{{Method}} & \multirow{2}{*}{{Backbones}}  & \multirow{2}{*}{{\#Frames}}  & \multicolumn{2}{c}{{Sth-Sth V1}}  & \multicolumn{2}{c}{{Sth-Sth V2}} & \multicolumn{2}{c}{{Jester}} & Throughput\\ 
    &&& \footnotesize{{\ \ Top-1 Acc.}} & \footnotesize{{GFLOPs}} &  \footnotesize{{Top-1 Acc.}} & \footnotesize{{GFLOPs}}  &  \footnotesize{{Top-1 Acc.}} & \footnotesize{{GFLOPs}}  & \scriptsize{(NVIDIA 3090, bs=128)}\\  
    \midrule
    TSN \cite{wang2016temporal} & R50 & 8 & 19.7\% & 33.2 &27.8\% &  33.2   & 82.6\% &  33.2 & - \\ 
    AR-Net \cite{meng2020ar} & MN2 + R18/34/50 & 8 & 18.9\% & 41.4 & -  &  -  & 87.8\% &  21.2 & - \\ 
    TRN\textsubscript{RGB/Flow} \cite{zhou2018temporal} & BN-Inc. & 8/8  & 42.0\% & 32.0 & 55.5\% & 32.0  & - & - & - \\
    ECO \cite{zolfaghari2018eco} & BN-Inc. + 3DR18 & 8 & 39.6\% & 32.0 & - & - & - &  - & -\\
    TANet \cite{liu2021tam} & R50 & 8 & 47.3\% & 33.0 &  60.5\% & 33.0  & - &  - & - \\
    STM \cite{jiang2019stm} & R50 & 8 & 47.5\% & 33.3 &  - & -   & -&  - & - \\
    TEA \cite{li2020tea} & R50 & 8 & 48.9\% & 35.0 &  60.9\% & 35.0   & -&  - & - \\
    TSM \cite{lin2019tsm}  & R50 & 8 & 46.1\% & 32.7 &  59.1\% & 32.7 & 96.0\% & 32.7 &  \ \ \ \ {162.7 Videos/s} \\
    AdaFuse-TSM \cite{meng2021adafuse} & R50 & 8 & 46.8\% & 31.5 & 59.8\% & 31.3   & -&  - & - \\
    \midrule
    TSM+ \cite{lin2019tsm}  & MN2 + R50 & 8+8  & 47.0\% & 35.1 & 59.6\% & 35.1 & 96.2\% & 35.1 &  \ \ \ \ { 123.0 Videos/s}   \\
    AdaFocusV2-TSM (128$^2$)  & MN2 + R50 & 8+12 & 47.0\% & \textbf{18.5} (\textcolor{blue}{$\downarrow$1.90x}) & 59.6\%  & \textbf{18.5} (\textcolor{blue}{$\downarrow$1.90x}) &  96.6\%  & \textbf{18.5} (\textcolor{blue}{$\downarrow$1.90x}) &  \ \ \ \ \textbf{ 197.0 Videos/s} (\textcolor{blue}{$\uparrow$1.60x})   \\
    AdaFocusV2-TSM (144$^2$)  & MN2 + R50 & 8+12 & 48.2\% & 23.5 & 60.5\% &  23.5 & - & - &  \ \ \ \ {159.6 Videos/s}   \\
    AdaFocusV2-TSM (160$^2$)  & MN2 + R50 & 8+12 & 48.6\% & 27.5 & 60.8\% &  27.5 & - & - &  \ \ \ \ {143.6 Videos/s}   \\
    AdaFocusV2-TSM (176$^2$)  & MN2 + R50 & 8+12 & \textbf{49.6\%} (\textcolor{blue}{$\uparrow$2.6\%}) & 33.7 & \textbf{61.3\%} (\textcolor{blue}{$\uparrow$1.7\%}) &  33.7 & \textbf{96.9\%} (\textcolor{blue}{$\uparrow$0.7\%}) & 33.7 &  \ \ \ \ {123.5 Videos/s}   \\
    \bottomrule
    \end{tabular}}}
    \end{footnotesize}
    \vskip -0.1in
\end{table*}


\begin{figure*}[t]
    \begin{center}
    \begin{minipage}{1.37\columnwidth}
    \centering
    \begin{footnotesize}
    \captionof{table}{\textbf{Ablation study of the training techniques.} Three representative conditions with different backbones, datasets and varying patch sizes are considered. We report the results of AdaFocusV2-MN2/RN and AdaFocusV2-TSM on ActivityNet and Sth-Sth V1.}
    \label{tab:training_tech}
    \vskip -0.175in
    \setlength{\tabcolsep}{1mm}{
    \vspace{5pt}
    \renewcommand\arraystretch{1}
    \resizebox{\columnwidth}{!}{
    \begin{tabular}{cccc|cc|cc|cc}
    \toprule
    Auxiliary & Diversity &  Stop- & Frame sampling & \multicolumn{2}{|c|}{ActivityNet (mAP)} &  \multicolumn{4}{|c}{Sth-Sth V1 (Top-1 Acc.)} \\
    supervision & augmentation & gradient & policy from \cite{ghodrati2021frameexit} & 128$^2$ & ${\Delta}$ &  128$^2$ & ${\Delta}$ & 144$^2$ & ${\Delta}$ \\
    \midrule
    & &  & & 69.4\% & -- & 38.2\% & -- & 40.7\% & -- \\
    \cmark& &  & & 72.5\% & +3.1\% & 42.9\% & +4.7\% & 44.6\% & +3.9\%\\
    \cmark&\cmark &  & & 74.8\%& +2.3\% & 46.1\% & +3.2\%  & 47.6\% & +3.0\% \\
    \cmark&\cmark &\cmark  & & 78.7\%& +3.9\% & \textbf{47.0\%} & +0.9\% & \textbf{48.2\%} & +0.6\% \\
    \cmark&\cmark &\cmark  & \cmark & \textbf{79.0\%}& +0.3\% & -- & --  & -- & -- \\
    
    \bottomrule
    \end{tabular}}}
    \end{footnotesize}
        \end{minipage}    
    \hspace{0.005in}
    \begin{minipage}{0.7\columnwidth}
        \hspace{-0.05in}
        \includegraphics[width=1.015\columnwidth]{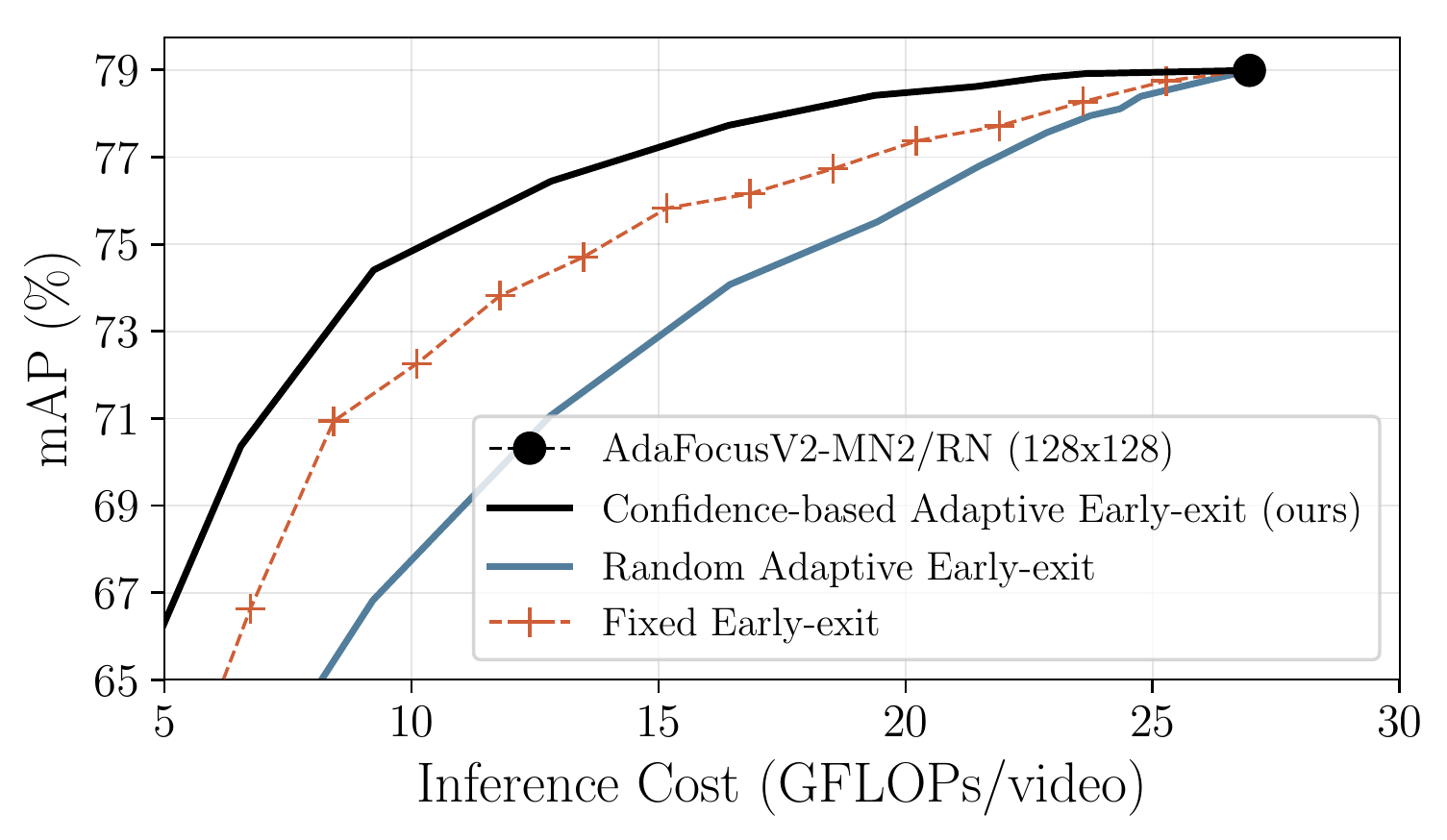}	
        \vskip -0.15in
        \caption{\textbf{Ablation study of AdaFocusV2+.}
    }\label{fig:abl_early_exit}
           
    \end{minipage}
    \end{center}
    \vskip -0.35in
 \end{figure*}


\subsection{Comparisons with State-of-the-art Baselines}

\textbf{Baselines.}
In this subsection, we compare AdaFocusV2 with a number of competitive baselines on ActivityNet, FCVID and Mini-Kinetics. AdaFocusV1 and several state-of-the-art frameworks for efficient video recognition are considered, including MultiAgent \cite{wu2019multi}, LiteEval \cite{wu2019liteeval}, SCSampler \cite{korbar2019scsampler}, ListenToLook \cite{gao2020listen}, AR-Net \cite{meng2020ar}, AdaFrame \cite{wu2019adaframe}, AdaFuse \cite{meng2021adafuse}, VideoIQ \cite{sun2021dynamic}, Dynamic-STE \cite{kim2021efficient} and FrameExit \cite{ghodrati2021frameexit}. An introduction is given in Appendix A.

\textbf{Implementation details.}
The comparisons are presented on top of the same backbone networks. Following AdaFocusV1, in most cases, we adopt MobileNet-V2 (MN2) \cite{sandler2018mobilenetv2} and ResNet-50 (RN) \cite{he2016deep} as the global encoder $f_{\textnormal{G}}$ and local encoder $f_{\textnormal{L}}$ of AdaFocusV2. When comparing with the methods that only leverage ResNet (i.e., FrameExit \cite{ghodrati2021frameexit}, Dynamic-STE \cite{kim2021efficient} and AdaFuse \cite{meng2021adafuse}), we replace the MobileNet-V2 with a ResNet-50 with down-sampled inputs ($96^2$). The two variants are referred to as AdaFocusV2-MN2/RN and AdaFocusV2-RN, respectively. A one-layer gated recurrent unit (GRU) \cite{cho-etal-2014-learning} with a hidden size of 2048 is used as the main body of the policy network $\pi$, while the accumulated max-pooling module \cite{ghodrati2021frameexit} is deployed as the classifier $f_{\textnormal{C}}$. The input frames are re-ordered following the sampling policy in \cite{ghodrati2021frameexit} (see Table \ref{tab:training_tech} for ablation). Due to spatial limitations, more details on network architecture and training hyper-parameters are deferred to Appendix B.

\textbf{AdaFocusV2 v.s. AdaFocusV1.}
The comparisons of AdaFocusV2 and AdaFocusV1 in terms of mean average precision (mAP) v.s. training/inference cost are presented in Table \ref{tab:actnet_vs_v1_train} and Figure \ref{fig:actnet_vs_sota}, respectively. Our method is implemented with the patch size of 96$^2$, 128$^2$, and 160$^2$. One can observe that AdaFocusV2 reduces the time consumption for training by $\sim\!2\times$, while dramatically improving the mAP using the same parch size (by 4-5\%). 

\textbf{Performance of AdaFocusV2+}
is depicted in Figures \ref{fig:actnet_vs_sota}, \ref{fig:actnet_vs_v1}. The curves correspond to reducing temporal redundancy on top of AdaFocusV2 (96$^2$, 128$^2$, 160$^2$), i.e., the three black dots. As stated in Section \ref{sec:v2_plus}, we vary the average computational budget, solve the confidence thresholds, and evaluate the corresponding validation accuracy. One can observe that AdaFocusV2+ effectively improves the inference efficiency of AdaFocusV2. In Table \ref{tab:actnet_vs_v1_plus}, we compare AdaFocusV2+ with AdaFocusV1+ by deploying them on top of the same AdaFocusV1 base model. AdaFocusV2+ achieves slightly better performance, and removes the requirement of additional reinforcement learning.

\textbf{Comparisons with state-of-the-art baselines}
on ActivityNet, FCVID and Mini-Kinetics are shown in Table \ref{tab:actnet_main_table}. It is clear that AdaFocusV2 (128$^2$) outperforms all the competitive efficient video recognition methods by large margins. For example, it achieves 4.2\% higher mAP (79.0\% v.s. 74.8\%) than VideoIQ \cite{sun2021dynamic}  on ActivityNet with similar GFLOPs. We further present the variants of the baselines with varying computational costs in Figure \ref{fig:actnet_vs_sota} for a comprehensive comparison. It can be observed that our method leads to a considerably better efficiency-accuracy trade-off. With the same mAP, the number of the required GFLOPs per video for AdaFocusV2+ is 1.9-3.6$\times$ less than the strongest baselines.


\subsection{Deploying on Top of Light-weighted Models}

\textbf{Setup.}
In this subsection, we implement AdaFocusV2 on top of a representative efficient network architecture, CNNs with temporal shift module (TSM) \cite{lin2019tsm}. We adopt the same network architectures and experimental protocols as AdaFocusV1. The MobileNet-V2 and ResNet-50 with TSM are used as $f_{\textnormal{G}}$ and $f_{\textnormal{L}}$. A fully-connected layer is deployed as $f_{\textnormal{C}}$ to average the frame-wise predictions as outputs. The policy network $\pi$ generates a single patch location for the whole video after aggregating the information of all frames, which is found important for high generalization performance \cite{Wang_2021_ICCV}. Training details can be found in Appendix B. Besides, for fair comparisons, the vanilla TSM is augmented by exploiting the same two backbone networks as ours (named as TSM+). TSM+ differentiates itself from AdaFocusV2 only in that it feeds the whole frames into ResNet-50, while we feed the selected image patches.

\begin{table}[t]
    \centering
    \begin{footnotesize}
    \caption{\textbf{Effectiveness of the learned patch selection policy.}}
    \label{tab:abl_policy}
    \vskip -0.175in
    \setlength{\tabcolsep}{1.5mm}{
    \vspace{5pt}
    \renewcommand\arraystretch{0.95}
    \resizebox{0.95\columnwidth}{!}{
    \begin{tabular}{c|ccccc}
    \toprule
    \multirow{3}{*}{\shortstack{Policy\\(128$^2$ patches)}} & \multicolumn{5}{|c}{ActivityNet mAP after processing $t$ frames} \\
    & \multicolumn{5}{|c}{ (i.e., corresponding to $\bm{p}_t$)} \\
    & $t$=1 & $t$=2 & $t$=4 & $t$=8 & $t$=16\\
    \midrule
    Random Policy & \ 37.7\% & 46.0\% & 56.8\% & 67.6\% & 73.9\% \\
    Central Policy  & \ 39.8\% & 47.7\% & 57.3\% & 66.6\% & 72.6\% \\
    Gaussian Policy  & \ 35.6\% & 44.5\% & 55.5\% & 67.1\% & 73.4\% \\
    \midrule
    AdaFocusV2-MN2/RN  & \ \textbf{44.8\%} & \textbf{52.5\%} & \textbf{61.7\%} & \textbf{71.0\%} & \textbf{76.4\%} \\
    \bottomrule
    \end{tabular}}}
    \end{footnotesize}
    \vskip -0.15in
\end{table}

\textbf{Comparisons of AdaFocusV2 and AdaFocusV1 on Sth-Sth}
in terms of training/inference efficiency are shown in Figures \ref{fig:fig1}, \ref{fig:sthv1_time_acc} and Figure \ref{fig:sth_flops_acc}. The end-to-end trainable AdaFocusV2 accelerates the training by 2.2-2.4$\times$, and improves the accuracy by 1-1.5\%/0.6-0.8\% on Sth-Sth V1/V2.

\textbf{Main results on Sth-Sth and Jester}
are presented in Table \ref{tab:sthsth}. With the reduced input size, AdaFocusV2 enables TSM to process more frames in the task-relevant region of each video, and effectively improves the inference efficiency. For example, AdaFocusV2-TSM reduces the computational cost of TSM+ by 1.9$\times$ without sacrificing accuracy. Notably, the practical speedup is significant as well.



\subsection{Analytical Results}

\textbf{Effectiveness of the proposed training techniques}
are validated in Table \ref{tab:training_tech}. One can observe that all of the three techniques significantly improve the performance of AdaFocusV2 across different experimental settings.

\textbf{Effectiveness of the learned patch selection policy}
is validated in Table \ref{tab:abl_policy}. Following AdaFocusV1 \cite{Wang_2021_ICCV}, here we do not reuse the global feature $\bm{e}^{\textnormal{G}}_{t}$ for recognition for a clean comparison. We assume that our AdaFocusV2 network processes a fixed number of frames for all videos, and report the corresponding mAP on ActivityNet. Three pre-defined policies are considered as baselines: (1) randomly sampling patches, (2) cropping patches from the centres of the frames, and (3) sampling patches from a standard gaussian distribution centred at the frame. One can observe that the learned policies have considerably better performance, especially when only processing parts of all frames.

\begin{figure}[t]
    \begin{center}
    \centerline{\includegraphics[width=\columnwidth]{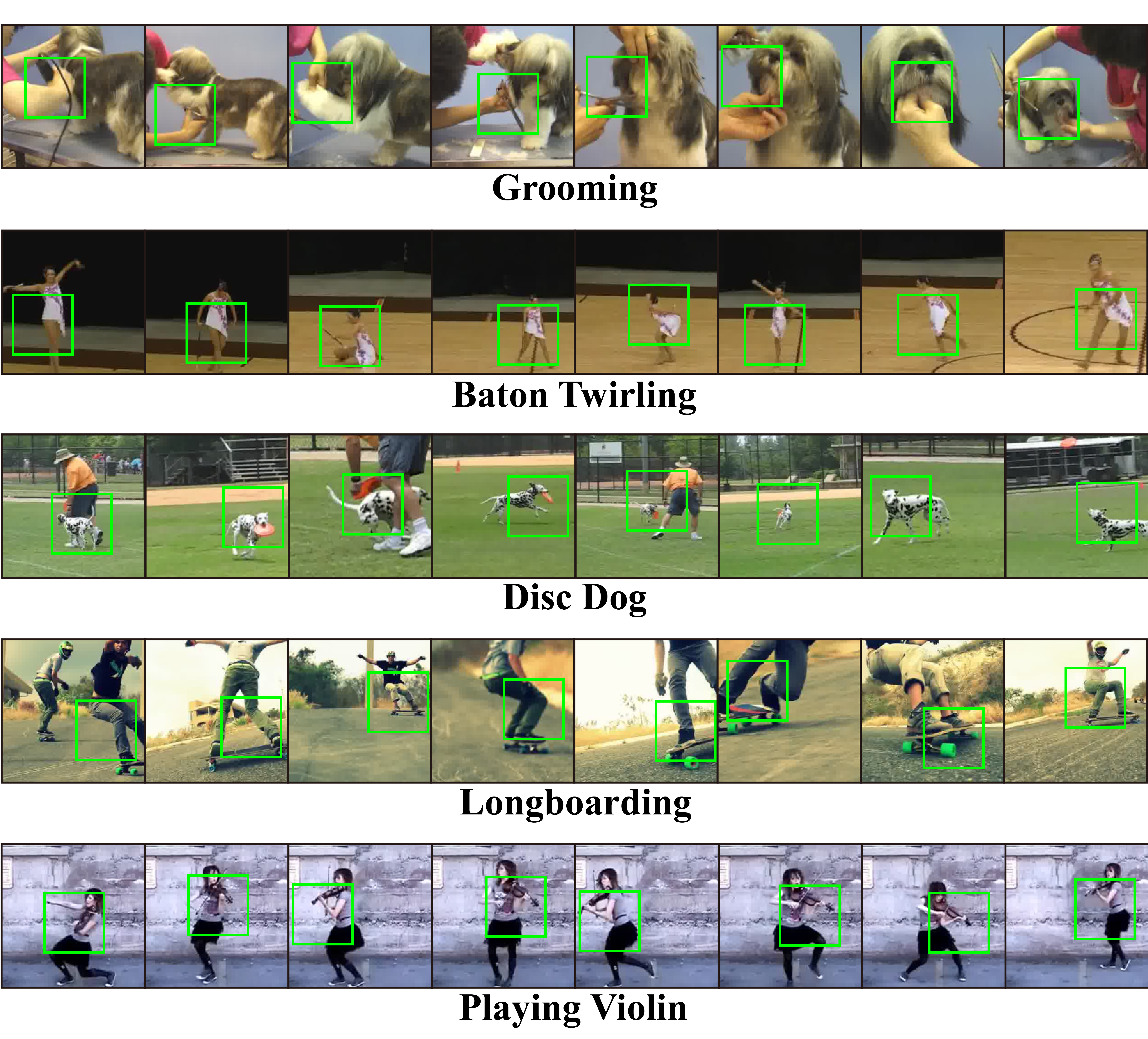}}
    \vskip -0.1in
    \caption{\textbf{Visualization results (zoom in for details).}  \label{fig:visualize}
    }
    \end{center}
    \vskip -0.4in
\end{figure}

\textbf{Visualization results}
are shown in Figure \ref{fig:visualize}, where the green boxes indicate the locations of the image patches selected by AdaFocusV2-MN2/RN (96$^2$). It is observed that the model attends to the task-relevant regions of each frame, such as the dog, the dancer, the skateboard and the violin.

\textbf{Ablation study of AdaFocusV2+}
is presented in Figure \ref{fig:abl_early_exit}. Two variants are considered: (1) random early-exit with the same exit proportion as AdaFocusV2+; (2) early-exit with fixed frame length. Our confidence-based adaptive early-exit mechanism outperforms both of them.




%% file: conclusion.tex
\vspace{-0.5ex}
\section{Conclusion}
\vspace{-0.5ex}

In this paper, we enabled the end-to-end training of adaptive focus video recognition networks (AdaFocus). We first proposed a differentiable interpolation-based operation for selecting patches, allowing the gradient back-propagation throughout the whole model. Then we present three tailored training techniques to address the optimization issues introduced by end-to-end training. Experimental results on six benchmarks demonstrated that our AdaFocusV2 network is considerably more efficient to train than the original AdaFocus model, while achieving state-of-the-art performance. 
